\documentclass{scrartcl}
\usepackage[utf8]{inputenc}
\usepackage[english]{babel}
\usepackage{color,soul}
\usepackage{marvosym}
\usepackage[affil-it]{authblk}
\usepackage{amsmath}
\usepackage{setspace}
\usepackage{amssymb}
\usepackage[shortlabels]{enumitem}
\usepackage{graphicx}
\usepackage[backend=biber, natbib, style=apa, citestyle=authoryear]{biblatex}
\DeclareLanguageMapping{english}{english-apa}
\begin{document}

\title{There is no Artificial General Intelligence}


\author{Jobst Landgrebe
  \thanks{Electronic address: \texttt{jobst.landgrebe@cognotekt.com}; Corresponding author} }
\affil{Cognotekt GmbH, Bonner Str. 209, D-50996 K\"oln, Germany}

\author{Barry Smith
  \thanks{Electronic address: \texttt{phismith@buffalo.edu}}}
			\affil{University at Buffalo, Buffalo, NY, USA} 

\date{\today}

\maketitle

\begin{abstract}
The goal of creating Artificial General Intelligence (AGI) --
or in other words of creating Turing machines (modern computers) that can behave in a way 
that mimics human intelligence -- has occupied AI researchers ever since the idea of AI was 
first proposed. One common theme in these discussions is the thesis that the ability of a 
machine to conduct convincing dialogues with human beings can serve as at least a sufficient 
criterion of AGI. We argue that this very ability should be accepted also as a necessary 
condition of AGI, and we provide a description of the nature of human dialogue in particular 
and of human language in general against this background. We then argue that it is for 
mathematical reasons impossible to program a machine in such a way that it could master 
human dialogue behaviour in its full generality. This is (1) because there are no traditional 
explicitly designed mathematical models that could be used as a starting point for creating such programs; 
and (2) because even the sorts of automated models generated by using machine learning, which 
have been used successfully in areas such as machine translation, cannot be extended to cope 
with human dialogue. If this is so, then we can conclude that a Turing 
machine also cannot possess AGI, because it fails to fulfil a necessary condition thereof. At the 
same time, however, we acknowledge the potential of Turing machines to master dialogue 
behaviour in highly restricted contexts, where what is called ``narrow'' AI can still be of 
considerable utility. 
\end{abstract} 

\section{Introduction}

Since the research field of AI was first conceived in the late 1940s, the ability of machines to 
conduct convincing dialogues with human beings has been seen as a necessary criterion for 
achieving artificial intelligence \citep{turing:1950}. There are important proponents of the 
attempt to realize what is called artificial general intelligence (AGI) who hold that the ability to 
engage in dialogue is ``only a sufficient, but not a necessary criterion for achieving AGI'' 
\citep{pennachin:2007}. Thus they argue that ``general intelligence does not necessarily 
require the accurate simulation of human intelligence'' \citep[21]{pennachin:2007}. 
There are good reasons, however, to take human intelligence as our starting point for 
understanding what AGI is. This is because, as the just quoted paper by Pennachin and 
Goertzel makes clear, when we engage in speculation as to the nature of `intelligence' 
independently of what we know about human intelligence, then we descend very quickly into a 
modern form of speculative metaphysics. Thus for example the authors define intelligence as 
the ``ability to perform complex goals in complex environments.'' According to this definition, 
the 1 mm-long nematode \textit{Caenorhabditis elegans} with 302 neurons would be 
intelligent, because it can achieve sexual reproduction (a complex goal) in its natural habitat 
(such as a pile of compost, which is undoubtedly a complex environment). 

We will assume further that the ability to use language is not only a very good proxy for general 
\textit{human} intelligence but also that it should be seen as a necessary condition for the 
existence of general \textit{artificial} intelligence. Communication using language is an activity 
that differentiates \textit{homo sapiens} from all other species. It is through language that we 
express our consciousness of reality and our ability to think and act freely. Conversation is the 
foundation of human society and of human culture. We believe that it is for this reason that 
Turing selected language as the core of his proposed method, later to be called the ``Turing 
Test'', to determine whether machines can emulate the human ability to think.  Though for 
many reasons (not analysed here), the ability to pass the test as it was described by Turing 
himself in 1950 is not a useful criterion for AGI, Turing's core idea -- namely that we can gauge 
intelligence by gauging the ability of a machine to mimic human dialogue -- remains central to 
our argument. For it will follow from the assumption that the ability to engage in fluent 
dialogue with human beings is a necessary condition for the existence of AGI, that if we can 
show that this ability is not realizable in a machine then we can infer that AGI, too, is 
impossible.

\subsection{Why dialogue matters}\label{matters}

Why, then, the central role of dialogue? This is because dialogue ability is critical for any 
practical use that we might want to make of AGI, for example in a business enterprise or in 
government. \textit{How could we primarily use AGI, if we had it?} What type of work could 
AGI machines perform that could not be performed either by human beings or by machines 
possessing one or other of the sorts of narrow AI that we already have at our disposal? We 
consider the possible answers to this question under three headings: (a) mobile physical work; 
(b) intellectual work that does not involve engaging in dialogue; and (c) work primarily 
involving communication. 

(a) Machines (robots) possessing AGI could use their intelligence to move freely and interact 
with dynamically changing, highly complex environments. Even if the work they were doing 
was entirely physical, for example transporting goods or disposing of waste, they would still 
have to be able to react to many kinds of environmental signals, among which human 
utterances are the most important. The utility of such machines would thus greatly increase 
were they able to understand and follow instructions issued by humans, even if they could only 
respond with stereotypical utterances such as ``Yes, master'' or ``I am sorry, master.'' Also, 
they should be able to react to human warnings, or understand suggestions from humans 
concerning better ways to do things. In other words, the ability to correctly interpret complex 
human utterances would still be required

(b) Machines performing intellectual work of the sort that does not involve spoken dialogue 
with humans -- for example loan application or insurance claim processing -- would still need 
to understand text, because the material they have to process is often provided in this form. 
Such work does not, in the normal case, require dialogue. But to process such documents with 
an error rate no worse than that of human beings, machines would need to exactly understand 
the meanings of the texts they process.

(c) Machines performing activities involving communication with human beings -- for 
example IM-chat or phone dialogues -- would need to be able to conduct such dialogues in a 
way that, at a minimum, allows the human user of the system to achieve her goals in an 
effective and efficient manner. And in order to justify a claim that a machine engaged in such 
activities possessed AGI, we would need to show that the machine has the ability to engage in 
dialogue with a human about an open-ended variety of topics in a way that does not require 
the human being to make specific sorts of extra efforts because they are dealing with a 
machine.

We note that in all three sorts of cases the AGI involved would need to demonstrate in its use of 
language to communicate with humans the ability to take account of highly complex contextual 
dependencies.

\subsection{AI conversation emulation and its failures} 

Many in the AI community are convinced that it is possible to create a machine with dialogue 
ability because they share Turing's view that building such a machine is just a matter of storage 
and computation  \citep{turing:1950, goertzel:2007}.  This in turn reflects a common 
assumption that human cognition and consciousness are themselves just a matter of storage 
and computation.

Here, however, we are not interested in the question whether machines can achieve 
consciousness, but rather only whether they can interpret complex texts and engage in 
dialogue with humans. Such conversation machines have been under construction since the 
1960s. Efforts are directed mainly towards what are called \textit{dialogue systems}, or in 
other words systems able to engage in two-party conversations, which are optimistically 
projected to be widely used in commercial agent-based applications in areas such as travel 
booking or service scheduling. However, despite major efforts -- from ELIZA 
\citep{weizenbaum:1966} to the computer-driven dialogue systems of the present day 
(including Siri and Alexa) -- nothing close to dialogue emulation has thus far been 
achieved.\footnote{See \ref{chat} below.} 

The tenacious optimism in the field is, we hold, based on the one hand on an unrealistically 
simplified view of what human dialogue behaviour involves, and on the other hand on a series 
of impressive successes in other areas of AI research -- above all in reinforcement learning, 
where solving the game of Go \citep{silver:2016} and achieving mastery in first-person shooter 
games such as Doom and Counter-Strike \citep{jaderberg:2018} have significantly raised 
expectations as to what might be possible in other areas. 

Counting against this optimism, however, is the repeated failure of attempts to build machines 
able to perform in a satisfactory way when engaging in dialogue with humans. This rests, we 
believe, on the complexity of the systems for generating and interpreting language that have 
evolved in humans and on the huge landscape of variance in natural language usage ensuing 
therefrom, some features of which were recognized by philosophers starting as early as Thomas 
Reid\footnote{\citet{schuhmannSmith:1990} provide a list of the variety of uses of language 
discussed by Reid, incorporating questioning (asking for information, for advice asking for a 
favour). providing testimony, commanding, promising, accepting / refusing (advice, a favour, 
testimony, a promise), contracting, threatening, supplicating, bargaining, declaring, and not 
last plighting (one's faith, one's veracity, one's fidelity.}, later by Schopenhauer\footnote{``It 
is by the help of language alone that reason accomplishes its most important achievements, -- 
the united action of several individuals, the planned cooperation of many thousands, 
civilisation, the state; also science, the storing up of experience, the uniting of common 
properties in one concept, the communication of truth, the spread of error, thoughts and 
poems, dogmas and superstitions.'' \textit{The World as Will and Representation}, \S 8.}, and 
then by Adolf Reinach\footnote{As \citet[1-2]{mulligan:1987} makes clear, the primary 
objective of both Reinach and his Anglo-Saxon successor J. L. Austin \citep{austin:1962}, ``is to bring into focus, and fully describe, a
phenomenon of which promising is their favourite example. Other social acts
dealt with in some detail by Reinach are requesting, questioning, ordering,
imparting information, accepting a promise and legal enactment, which --
except for the last two -- are all at least touched on by Austin. In all
these social acts we have `acts of the mind' which do not have in words and
the like their accidental additional expression. Rather, they `are
performed in the very act of speaking'. These cases of doing something by
saying something are, and give rise to, changes in the world. They are
associated with a variety of different effects. Examples of the effectivity
(Wirksamkeit) of social acts are both the obligations and claims to which
promises and orders give rise and the behaviour, whether a social act or a
non-linguistic action, which some social acts are intended to bring about.''}.

Arnold Gehlen explored the field from the biological and anthropological perspective in his 
main work \textit{Man}, first published in 1940 \citep{gehlen:1988}. A decade later 
Wittgenstein's \textit{Philosophical Investigations} gave rise to a significant enhancement in 
our understanding of how (especially spoken) language works, which was refined by the 
contributions of philosophers such as Austin and Grice and has since been consolidated in the 
huge body of research in the philosophy of language, linguistics, semantics and pragmatics, 
upon which we draw extensively in what follows.

\subsection{Main arguments of this paper}

To see why human dialogue eludes mathematical modelling
we must first describe how humans generate and interpret language when engaged in dialogue 
and why the capacity for such generation and interpretation is part of an essential survival 
strategy for \textit{homo sapiens}. We shall see that language is a sensory-motor human 
capability, which arose in the evolution of our species as a genus-specific way to interact with 
our environment in an abstract, controlled fashion that replaces the instinctive behaviour of 
our non-human ancestors \citep{gehlen:1988}. Briefly, language enables us to shape and 
realize our intentions (for example through deliberation and planning), and we will show that 
this implies a potentially infinite variance in the ways we use it. We contrast this to the 
capabilities of Turing machines,  first expressed more than 175 years ago by Ada Lovelace in 
her statement to the effect that the Analytical Engine\footnote{This is the machine built by 
Charles Babbage which, as Turing points out (\textit{op.cit.}), is mathematically equivalent to 
a Turing machine.} ``has no pretensions to originate anything. It can do whatever we know 
how to order it to perform'' \citep{lovelace:1842}. We interpret Lovelace here as asserting that 
the machine will never develop a counterpart of (for example) human
intentions\footnote{This is true even though dNN can now be built that
develop new automated models which have not been designed explicitly by
humans, see section \ref{stochPM}.}, and thus also 
not learn in the way that humans do, \textit{until we know how to tell it to do so}. But to 
achieve this, we would have to create mathematical models of these human characteristics, 
since -- as was often pointed out by Turing himself -- \textit{we can only model what we can 
describe mathematically}, and we shall see that this is beyond the bounds of what is possible 
given our current mathematics. We thus go one step further than Searle's Chinese room 
\citep{searle:1980}, which states that machines cannot emulate consciousness: For we reject 
the very idea that a Turing machine could be built that would emulate human conversational 
behaviour.

\subsubsection{Mathematical models of human dialogue}

What, then, of the mathematical representation of human dialogue? A dialogue is a complex 
stochastic temporal process of a certain sort -- as we shall see, it is a process that lacks the 
Markov property (according to which state transition probability depends only on the 
immediately preceding state). Processes in the human brain quite generally are of this sort, as 
are (for example) the processes generated by the global climate system. All such systems, as we 
shall see, elude mathematical modelling.

Certainly, some stochastic processes can be modelled mathematically using what are called 
stochastic models \citep{parzen:2015}. But for this to be possible, we need input-output data 
tuples where the inputs are connected to the outputs probabilistically, which means that there 
is a certain (measurable) likelihood that a given input will be associated with a given output 
\citep{landgrebeSmith:2019}.

This is the basis of so-called ``machine learning'', which applies in the most straightforward 
case in situations in which (1) human beings repeatedly process data in a certain way, (2) we 
are able to collect large quantities of the input data that are used for this purpose and (3) 
associate these data with the output data humans have created therefrom. An example is the 
behaviour of humans in identifying spam in their email.  Here the sender, subject and text of 
an email serve as input, while the human decision to put this email into the spam folder 
provides the output. The process of training the machine with these data yields a gigantically 
large equation that models the relationship between the input and output data that have been 
used for training. This model is then narrowly tied to the training data that generated it, so that 
if the equation (the trained model) is applied to new data that is not drawn from the same 
distribution as the original training data, it will compute undesired outputs.\footnote{We deal 
with this matter at greater length in our discussion of the spam filter and other examples in 
\citep{landgrebeSmith:2019}.}

Only if we have a sufficiently large collection of input-output tuples in which the outputs have 
been appropriately tagged can we use the data to train a machine that, given new inputs 
sufficiently similar to those in the training data, is able to predict corresponding outputs with a
certain (useful) degree of accuracy.

What `sufficiently large' and `sufficiently similar' mean, here, are questions of mathematics. 
We shall see that, when these questions are raised in relation to those sorts of stochastic 
temporal processes which are human dialogues, then it becomes clear that there are in fact 
\textit{three} insurmountable hurdles to realizing the scenario in which a machine could be 
trained to engage convincingly in human conversation, namely

\begin{enumerate}
 \item that human dialogue processes do not meet the conditions needed for the application of 
any known type of mathematical model,
\item that, due to the inexhaustible variance which human dialogues exhibit -- which is as 
huge as the variance in human culture and behaviour in its entirety -- we could never have 
sufficiently large amounts of data to train a machine, and
\item that to learn the correct interpretations of the dialogue utterances, interpretations which 
are indispensable to adequate dialogue production, the utterances themselves are insufficient; 
what is implicit in the dialogue cannot be fully derived from what is given explicitly.
\end{enumerate}

We shall see that nothing has changed in this respect even with all the advances made in 
machine learning in recent years (including reinforcement learning (see \ref{rl}), adversarial 
learning and unsupervised sequence learning (see \ref{mp})). For again: the limitations on 
what machine learning can do are of a mathematical nature. 

Nothing has changed, either, as a result of the impressive accumulation of data pertaining to 
human language use, including large amounts of dialogue content in the form of Youtube 
interviews or of data deriving from use of Siri, Alexa and similar services. Interview data are 
typically highly stylized and thus represent only a small fraction of the variance needed to be 
useful to support machine learning of the sort needed to implement a machine counterpart of 
general human dialogue. Siri/Alexa data, on the other hand, are merely recordings of human-
machine interactions, and so are of zero utility for our purposes here. Both sorts of data are 
also severely limited by the restriction to what is explicit, and thus recordable. 

\subsubsection{There can be no AGI}

In this communication, we will give evidence for the soundness and validity of the following 
syllogism concerning the creation of AGI:

\begin{itemize}
 \item There can be no mathematical models for the type of behaviour occurring in human 
dialogues. 
 \item Therefore, there can be no computer programs implementing such models. 
\item The ability to emulate human dialogue behaviour is a necessary condition for 
implementing AGI.
 \item Therefore, there can be no AGI.
\end{itemize}

This does not of course mean that all is lost for AI in the realm of human dialogue. For we also 
review the current state of the art in dialogue system building and conclude by identifying what 
we see as the potential for dialogue systems that would still be \textit{useful} even though they 
fall far short of AGI. This essay thus complements our previous paper 
\citet{landgrebeSmith:2019}, where we defended a sceptical attitude to the current euphoria 
surrounding ``deep neural networks'', while at the same time pointing to AI applications 
which provide significant utility in addressing specific sets of real-world problems.

\section{Language and dialogues}

\subsection{The nature of human language}\label{natLang}

Our use of language is the expression of our will to interact with the physical world around us 
and with other humans in pursuit of our goals \citep{schopenhauer:1819}. We shall refer to 
these goals as they impact our day-to-day behaviour as \textit{intentions} 
\citep{bratman:2009}. 

We concentrate in what follows on the use of spoken language, which is a sensory-motor 
activity closely related to, for example, hand movements involved in grasping an 
object.\footnote{These similarities were first comprehensively described by 
\citet[chapters~19ff.]{gehlen:1988}; for a contemporary treatment, see \citet{gomez:2013}.} 
When we perform motor activities, we simultaneously obtain propriosensory feedback from 
the performance itself. In the case of the arm-hand movement for grasping, proprioception is 
augmented by a second sort of feedback deriving from the object as we touch it, feedback which 
confirms that we have achieved our grasping intentions. These two sorts of feedback -- from 
our own body and from our environment via our sensory system -- allow us to continuously 
adjust our intentions. 

Language is however a much more powerful type of motor activity than all the others. The 
sensory-motor feedback occurs as we hear our own words as we are speaking; but now the 
proprioception is augmented through the feedback we receive from our interlocutor -- for 
example in the form of facial expressions, gestures, as well as further speech. We continuously 
use this feedback to adjust not just what we say and the way we speak but also the intentions 
we are seeking to realize by engaging in dialogue \citep[chapter~33]{gehlen:1988}.

\subsubsection{Language functions}

Animals act on the basis of instinct, and they therefore ignore those sensory inputs that do not 
stimulate their pre-defined response spectrum. A passive, static filter blocks non-relevant 
stimuli and lets through only those stimuli that trigger instinctive behaviours 
(\citet[chapter~15]{gehlen:1988}, \citet{milton:2000}).

Humans, in contrast, have no such passive filter governing what they experience, but are able 
to deal freely with the full breadth of sensory inputs. In contrast to animals, humans live in this 
sense in an open world \citep{scheler:1976}. This is, on the one hand, a defect. For it means 
that humans do not have at their disposal the sorts of instinctive routines that would make 
them well adapted to their natural environments. But on the other hand it is a benefit, since it 
means that humans are adaptable to ever new environments through use not only of their 
mental capacities but also of tools (including language). This adaptability is seemingly without 
limits. Hence \textit{general intelligence}.

At any given moment, humans can choose from a broad and ever-changing repertoire of 
environmental inputs, and already from early infancy humans manifest strategies to avoid 
being overwhelmed by sensory inputs. These include apprehending their environment not as a 
meaningless mosaic of sensations, but rather as a world of enduring objects (including 
persons) divided into different kinds (for example animate and inanimate), linked by causal 
relations, and manifesting characteristic functions and rule-governed behaviours 
\citep{gibson:1979, keil:1989}. 

Over time, infants apprehend a subset of the behaviours of their fellow human beings as of 
special significance, namely those which involve the production of linguistic utterances, which 
they themselves imitate. With increasing sophistication, they themselves begin to use language 
in ways which enhance inborn tendencies to recognize classificatory hierarchy and causal and 
functional patterns in the world. For example, they learn to identify a wooden stick as a pencil, 
and thus as a tool for writing. By using general terms to describe both the things in the world to 
which our intentionality is directed and the associated sensations and emotions, language 
enables us to distance ourselves from our immediate experience of what is particular in 
external reality and from our spontaneous emotional reactions 
\citep[chapter~28]{gehlen:1988}. 

In his paper about illocutionary acts, \citet{searle:1975} introduces the distinction between two 
`directions of fit' between words and world. On the one hand is world-to-word direction of fit, 
for example when I make a list of items I need to pack for the holidays and then act in such a 
way as to make the world match my list. On the other hand is the word-to-world direction of 
fit, when I make a list of items actually packed. Lists of this sort illustrate how language creates 
a new plane of activity through which we can shape and view the world. I can use my list as a 
tool to help me realize my intention \textit{to pack my bag}, or as a basis for reflecting on what 
items I will really need. 

The list illustrates more generally how by using language (including silent soliloquy) we are 
able to distance ourselves from the stream of our present experiences, both inner and outer. 
The example list shows us how we use language to engage in \textit{planning behaviour}, 
which involves engaging in a sort of abstract simulation of different courses of action. The 
example also illustrates the way in which language serves as the foundation of progressively 
more ambitious social interactions: from using a simple list as the means to have someone else 
pack your bag to enabling the sorts of collective agency needed to build cathedrals and space 
ships or maintain a legal system or an industrial enterprise.

Language allows us to deal with increasing sophistication with patterns in the ways causal 
(including intentional) processes unfold and to shape these patterns by providing a powerful 
vehicle for the forming and realization of our individual and collective intentions. It allows us 
to react in ever more flexible and useful ways to what would otherwise be an overwhelming 
flood of stimuli, and to do this at successively more general and abstract levels as concerns not 
only our interactions with the external world but also the ways we cope with our own inner 
sensations and emotional reactions.\footnote{As when someone reasons in their mind as 
follows: ``I have been feeling angry in situations like this before, but my previous overreaction 
only aggravated my situation, so this time I will attempt to control my responses.''}

\subsubsection{Foundations of the language background}

Language is a complex of capabilities that are applied by humans to enhance their processing 
of both external and internal reality. Although the ability to use language is genetically 
encoded, the specific languages that people use are parts of culture and must be learned. Each 
individual uses language in a specific way, which depends on their specific experiences in 
interacting with the world (including their fellow human beings) from infancy onwards. 

In particular, the general terms we use in describing reality have a foundation in our physical 
experience. We learn to use `bitter' by registering that contexts in which we experience tastes 
of a certain sort go hand in hand with contexts in which people use this word. Such 
abstractions can also arise from associations at higher levels, as when the positive feeling we 
experience when eating something sweet gives us an understanding of the adjective `delicious'. 

\subsection{Dialogues}

We engage in dialogues in order to interact with other people to achieve certain ends (berating, 
guiding, learning, persuading, socializing, and many more). As our interlocutor responds, we 
take what we hear and view it, typically spontaneously and unconsciously, in light of our 
current intentions and what we have experienced in previous encounters. We thereby once 
again condense the sensory input down to the abstract linguistic plane encompassing just what 
is needed to understand what has been said. 

Utterances and interpretations take place in time and (more or less) in sequence. Both involve 
the making of conscious and unconscious choices, which are \textit{implicit} in the sense that 
they are accessible to the dialogue partner -- and to any external human or machine
observer -- at best indirectly, for example via facial expressions or via the utterances to which 
they lead. 

Our intentions thereby interact with the intentions of our interlocutor as the dialogue proceeds 
through successive cycles of turn-taking,\footnote{Turn taking is guided by rules and also by 
what (Sacks\textit{op. cit.}) calls turn-constructional units, an important subtype of
which are ``possible completion points'', which are signals in the dialogue that indicate the 
opportunity for a role switch.} a phenomenon which seems to be found in all human
cultures \citep{schegloff:2017, stivers:2009}. In each cycle the drivers of the conversation for 
both participants are their respective intentions -- the goals they each want to achieve by 
means of their utterances \citep{grice:1957, austin:1962, searle:1983}.\footnote{If
the dialogue arises spontaneously, only the first utterer may have an intention; but the 
interpreter will very quickly form intentions of her own as soon as she is addressed, including 
the intention to refuse engagement in a dialogue.} When it is Mary's turn to speak in a dialogue 
with Jack, she tries to fulfil her intentions by conveying content meaningful to Jack and in a 
way that Jack will find persuasive. As Mary tries to influence Jack, so she in turn may be 
influenced by the ways in which Jack responds. In this way, a conversation will typically bring 
about changes in the intentions of its participants. A speaker may foresee the reactions of his 
interlocutor to his utterances and consciously or unconsciously plan out the conversation flow 
in advance. Creating such a plan is sometimes even the explicit intent of the conversation, as 
when people sit down together to reach decisions about how to synchronize their intentions.

\subsection{Habits, capabilities and intentions}\label{ident}

In every case, dialogue interaction take place against an enduring, and typically slowly 
changing, background, consisting of the evolving intentions of the interlocutors and of their 
respective personalities, habits, capabilities and other elements drawn from their personal
biographies. 

These form what we shall call the `identity' of a human being, by which we mean that highly 
complex individual pattern of dispositions, among which the most important are (in ascending 
order) the visceral, motor, affective and cognitive dispositions that determine a person's 
possibilities of reaction to internal or external stimuli.\footnote{The underlying account of 
dispositions is sketched in \citet{hastings:2011}. This draws in turn on
the ontology framework described in \citet{arp:2015}}

Your identity, in this sense, results from the combination of genotypic and
environmental influences which affect your neural (or more generally your 
physical\footnote{In language production the physical substrate consists not only of your  
brain but also of your diaphragm, lungs, and the entire vocal apparatus.}) 
substrate as it develops
through time.\footnote{Our `identity' thus comes close to what Searle calls
`The Background' \citep{searle:1978}, of which Searle himself says that it is at
one and the same time (i) ``derived from the entire congeries of relations
which each biological-social being has to the world around itself'' and
(ii) purely a matter of that being's neurophysiology \citep[154]{searle:1983}.

We can distinguish three main families of dispositions through the
realization of which our identity is manifested:

\begin{enumerate}
\item  habits, tendencies, personality traits (for example tendencies to stutter, to fret, to avoid 
commitment, to behave politely, to behave honestly, $\dots$)
\item capabilities (to speak a language, to play the piano, to manage complex activities, to do 
long division, to play championship tennis, to practice law, $\dots$)
\item intentions, goals, objectives (to pass this or that exam, to marry Jack, to impress Jack's 
mother, to lose weight, to heal the rift with your bother, $\dots$)
\end{enumerate}

Our intentions are the drivers of our behaviour and are typically short-lived; intentions may be 
adjusted, for example, with each successive utterance in a dialogue. Our habits and capabilities 
are longer lasting. They rest on enduring patterns in the underlying physical substrate and 
shape which intentions we develop and how (and whether) they are realized. 

In spite of all the advances in neurology in recent years, the human neural substrate is still 
little understood. Indeed, it is not understood at all if we define `understanding' as the ability 
to model and predict the phenomenon we claim to understand. Thus, it
cannot be captured in a formal, let alone machine-processable, way. 
The same applies also to the array of dispositions -- which we share to a greater
or lesser extent with our fellow human beings -- of which it serves as the material basis. It is 
this array of dispositions that makes conversation (and indeed
all use of language, indeed all human activity) possible. It shapes and determines the repertoire 
of the types of speech acts that we have at our disposal while at the same time ensuring that the 
deployment of this repertoire is to a large extent a matter of ingrained reflex -- or at least a 
matter over which we have only very fragmentary conscious control \citep{billig:1997}.

Realizations of our linguistic dispositions are triggered in various ways,
including by the utterances of our dialogue partners. Sometimes, such
realizations may involve conscious choices, for instance the choice of whether
to adopt a retaliatory or conciliatory tone in response to a threatening
utterance, or the choice of which answer to give to a difficult (perhaps a
trick) question. More often, however, selection takes place spontaneously
and unconsciously. It occurs, moreover, on a number of different levels,
affecting both verbal and non-verbal aspects of communication, and we document in the 
Appendix } \ref{vars} the huge variance involved in the different sorts and features of 
utterance structures that can be produced in the course of a dialogue.

Matters are made still more complicated by the fact that a decisive role in the formation of both
utterances and interpretations is played by the contexts in which communicative acts take 
place \citep{fetzer:2017}. We show in the Appendix \ref{context} that there is a vast range of 
multiple types and levels of such contexts. And, to make matters worse, the range of possible 
choices is not static or stable \citep[59]{verschueren:1999}. 

The result is that there are so many different sources and dimensions of variance involved in a 
communicative act that the possibilities of forming an utterance are practically infinite. 
Humans can cope with this degree of variance because they can actively form and interpret 
utterances based on their own intentions. Even a total lack of understanding of a sentence 
spoken in a foreign language can be brought into congruence with one's own intentions, for 
example by actively giving up the attempt at communication or by communicating using 
gestures. 

\section{Why machines cannot conduct real dialogues} \label{can1}

For a machine to possess dialogue ability, it would have to display the same ``general
experiential understanding of its environments that humans possess'' 
\citep{muehlhauser:2013} and also the same spectrum of abilities to react to these 
environments that humans possess (including human reactions that fall short because they 
involve mistaken uses of language, or errors resting on misunderstandings, or slurring of 
words resting on intoxication, and many other departures from the norm). 

How a dialogue participant reacts at each moment of a dialogue is determined

\begin{itemize}
\item by his intentions of the moment,
\item by his language abilities,
\item by what he perceives in the course of the dialogue itself,
\item by what he (most of the time unconsciously or implicitly) remembers (both emotionally 
and intellectually) from his life experiences and
\item  by how all of these factors are related together.
\end{itemize}

\subsection{Language as a necessary condition for AGI: Criteria}\label{criteria} 

We think that mastering of language and dialogue is a necessary condition for AGI because it is 
the primary medium of expression of the human intellect and because many conceivable AGI 
applications would need to interact with humans via language. If, therefore, the following 
criteria could be satisfied by a machine engaging in spoken dialogue, then we believe that this 
would provide strong evidence for its being a realisation of AGI\footnote{These criteria form 
the basis for a more realistic version of the Turing test that is described in 
\textsc{Forthcoming}, Grazer Philosophische Studien.}: 

\begin{enumerate} 

 \item the machine has the capability to engage in a convincing manner with a human
interlocutor in dialogues of arbitrary length in such a way that the human interlocutor does not 
feel constrained in the realisation of his dialogue-triggering intentions by the machine-dialogue 
partner. This means that when the human interlocutor engages in the dialogue, he must be 
able to realise his intentions without making the sort of special effort he would need to make 
when dealing with a machine.

 \item the cycles in such a dialogue are not restricted to cases where the machine merely reacts 
to a human trigger, as in a succession of question-answer-pairs; rather, the interlocutors 
behave exactly as they would in a normal dialogue;

 \item that the dialogue would be in spoken form\footnote{A spoken dialogue of this sort 
would require a solution to the (hard) problem of engineering a machine with a
voice production capability that does not impede the dialogue flow to avoid a violation of the 
first criterion.} and

 \item that the machine would \textit{see} the human interlocutor, since the
machine has to demonstrate that it can react appropriately to the whole
habitus of its human dialogue partner and not just to her speech: Many utterances cannot be 
adequately interpreted without taking into account gestures and facial expressions. A machine 
without vision would thus not be able to perform the utterance interpretations expected in 
many types of human dialogue\footnote{Note that this would require that the machine learns 
to integrate visual sensory input into the interpretation of language. The problem of visual 
sensory input interpretation is harder than language interpretation because visual input is 
non-processed raw input devoid of any semantics. Animals interpret it based on instincts, but 
humans interpret it based on the meaning for their survival. Currently, machines do not 
interpret visual input at all: contrarywise, when they classify images, they use image elements 
that are unrelated to the elements  used by humans \citep{moosavi:2017, jo:2017}. Not only do 
we not know how to change this in dNN, but we have no way of modelling how our mind 
integrates sound and vision, when, for example, interpreting a slapstick-scene in a movie and 
laughing about it.}.

\end{enumerate}

\subsection{Human and machine identity}\label{machineBio}

When a normal human being engages in conversation, she is able to draw on her entire 
personal history and on her repertoire of capabilities, not just of a
linguistic nature, but also capabilities she has acquired in the course of her life in
navigating many other aspects of reality. She is able to manifest, in other words, what we have 
called her ``identity'' (see section \ref{ident}). The vividness and emotional adequacy of a 
dialogue requires an identity as dialogue foundation. Machines do not have personalities or 
identities. Therefore, a dialogue with a machine will always have a static character and lack the 
vivacity conveyed by the richness of a real life. Therefore, the first of the criteria we list above, 
namely that the human interlocutor is able to realise his intentions in the course of the 
dialogue without making the sort of special effort applied when dealing with a machine, will 
not be satisfied.

\subsection{Initial utterance production}

In providing an account of the powers that would be required of a machine
purporting to emulate human dialogue behaviour, we distinguish between two sorts of task: (1) 
the production of the initial utterance of a dialogue, and (2) the maintenance of subsequent 
dialogue flow. 

The act of producing an initial utterance requires only the ability to understand the context in 
which the dialogue partners find themselves, while dynamic dialogue maintenance requires 
taking into account the switching of roles over time. We will begin with the initial utterance, 
and show that the machine struggles even here. 

\subsubsection{Initial utterance production by machines}

Contexts of the sort in which an AGI might need to produce an initial utterance are, for 
example, a traffic accident, where the AGI acts as robot policeman or paramedic. The AGI 
would need to understand the situation in order to make an appropriate initial utterance. This 
is not by any means a trivial task, given the massive variation in real traffic situations we are 
faced with in everyday life.\footnote{The huge degree of variance can be understood by 
examining court rulings arising from random traffic disputes.} The AGI would need not only to 
understand the overall situation, but also to find the appropriate words to use when speaking 
to just these human beings in just this psychologically fraught situation. Pre-programmed 
initiating sequences, such as ``Hello, I am your automated police officer Hal. I have registered 
your participation in an accident. Please show your driver's license'' typically will not 
do\footnote{Matters of suitable tone, prosody and intonation would also have to be taken into 
account, see Appendix \ref{soundS}}.

\subsubsection{Initial utterance interpretation by humans} \label{human-int}

What now as regards the \textit{interpretation} of a single utterance of
the sort we are called upon to perform in relation to the first utterance in a dialogue? For 
humans, according to current understanding, this task
has two steps: first is a syntactic step, which is realized through a
dynamic process of syntactical sentence parsing and construction using the
structural elements constituting the uttered sentence.\footnote{There are several
grammatical theories about how this happens, ranging from generative to
constraint-based theories. \citet{mueller:2016} gives an overview.}

This syntactical analysis yields the basis for the second, semantic step,
which is the context-dependent assigning of meaning to the uttered sentence
\citep{loebner:2013}. Even for one sentence this process has a dynamic
aspect. This is because, beginning with the very first word, the syntactic
construction and semantic interpretation interact. This can require several
successive cycles of revision, as an initial syntactic construction is
revised as earlier parts of the sentence are re-interpreted in light of the
ways they interact with parts coming later. \citet{auer:2009} has
coined the term ``on-line syntax'' to describe this phenomenon.

For a single utterance in a face-to-face dialogue, the core context required for its interpretation 
by a human is what \citet{barker:1968} refers to as the ecological setting. This
is the salient part of the environmental (physical) context in which the
dialogue takes place and which will typically be centred on the person by
whom the initiating utterance is made. When a dialogue is initiated on the phone, the absence 
of such a context explains why many humans find it hard to speak with someone they have 
never met or spoken to: the absence of a shared physical environment severely reduces the 
amount of context usable by the interlocutors and thereby creates a barrier to the transmission 
of meaning.

When interpreting the single utterance, the human has to apply contexts available to her from 
her own biography together with any clues she can draw from her interlocutor's tone, dialect, 
physical appearance, behaviour and so forth. Discourse economy\footnote{Cf. Appendix 
\ref{impl}} forces her to make assumptions on this basis in her attempt to understand those 
aspects of meaning left implicit by a speaker, for example in order to disambiguate ambiguous 
aspects of his utterance, or gauge the force of turns of phrase that might in some contexts be 
threatening or indicative of deceit. In face-to-face conversations, humans can use contextual 
cues to achieve this. For example, when negotiating the purchase of a used car, the buyer will 
look for non-verbal cues indicating the reliability and honesty of the seller to make up for the 
information asymmetry inherent to the situation \citep{akerlof:1970}.

In addition, humans interpret static utterances by using knowledge they
have derived through processing their own experiences over time, above all
knowledge acquired through practical experience of the way the world around
them is structured causally. From these experiences (combined with innate
capabilities) they acquire an ability to reason about the relationships
which link together entities in their environment into different families of
predictable patterns.\footnote{This ability is sometimes called `common
sense' \citep{smith:1995}. Compare also section 2.2 of
\citet{landgrebeSmith:2019}.} The latter are then extended also to the
entities referred to in dialogue utterances, and this enables these
utterances to be interpreted, for example in terms of their practical
relevance to the interpreter.

\paragraph{Human interpretation of multi-sentence initial utterances}
\label{mssu} Initial utterances consisting of more than one sentence are
still more challenging for humans to interpret than single-sentence
utterances. This is because the sentences now contextualise each other:
there are syntactic and semantic as well as explicit and implicit
interdependencies which link them together.\footnote{We deal in Appendix \ref{disDeix}
with the phenomenon whereby a part of a dialogue can itself serve as context for another part 
of the dialogue.} For example, sentences may be
connected explicitly, via anaphora, or as chains of steps in an argument or
chronological narrative, or implicitly, through analogies or historical
resonances attached to certain words or phrases.

\subsubsection{Initial utterance interpretation by machines}

How, then, does the \textit{machine} interpret the initial 
utterance? Here again two steps are involved: of syntactic construction and
semantic interpretation. We deal with these in turn.

The syntactic construction using structural elements that humans perform
according to the grammatical theories referred to in 
\ref{human-int} can be mimicked by the machine quite effectively for
written text, when no non-lexematic structural language material has to be
taken into account.\footnote{By `lexematic material', here, we mean those
structural elements that can be directly reduced to lexemes -- essentially
wordforms and all their variants and composites. Lexematic material is subset of verbal 
material, while non-verbal material is the part of communication that is not produced via 
sound: facial expression, gesture, posture, movement patterns.}
Machines fail, however, as soon as non-\-lex\-ematic structural material such as facial 
expression, gestures, posture, or sound structures come into play (see Appendix \ref{soundS}). 
This is because the world knowledge enabling the interpretation of this material
-- which can be combined in arbitrary forms to create many different sorts
of contexts -- cannot be learned without life experience and it cannot be
mathematically formalised (see Appendix \ref{natLang} and sub-section
\ref{stt-nl} later in this section).

For the interpretation of a single sentence -- ignoring for now gestures and other non-
lexematic material -- the machine would need to reproduce the syntactic
construction achieved by humans if the static interpretation pattern used
by the human brain (syntactical analysis followed by semantic step) is to
be reproduced.\footnote{The core of the machine-learning-NLP community currently thinks
this is no longer necessary. All is supposed to be computed implicitly
using ``end-to-end deep neural networks''. See \citet{manning:2016}.}

This requires use of computational phrase structure grammar, dependency grammar or 
compositional grammar parsers.\footnote{An overview is given in \citet{manning:1999}.} 

All of these create trees which represent the syntactic structure of the sentence.
The parsers work well if the input sentences are syntactically valid.
However, if a sentence is syntactically valid but semantically ambiguous,
as in:

\begin{enumerate}
\item[(4)] He saw old men and women,
\end{enumerate}

\noindent an ideal computational parser will create two syntactic trees
representing each sense.\footnote{This feature is only available with
compositional grammar parsers \citep{moortgat:1997}. With a sufficiently
sophisticated computational setup, a context-dependent disambiguation may
be possible.}

It is with the \textit{interpretation} of the syntactic structure -- in
other words with the move from syntax to semantics -- that machines
struggle, and this holds even in the static single sentence utterance case.
\textit{For what is the context which the machine could use to assign
meaning to a single sentence?} 

The machine cannot decide this on its own. The multitude of combinations of language 
elements described in Appendix \ref{vars} 
allows allow for a huge number of interpretation possibilities even at
the single sentence level. The machine cannot decide, for instance, how to
fill in implicit meaning generated as a result of language
economy, or of the use of incomplete utterances or ellipses.

To achieve this, the machine would need an appropriate context and dialogue
horizon. Background information would thus need once more to be
\textit{given} to the machine, analogous to the sort of information given to an undercover 
agent to provide him with a cover story -- information needed to enable the machine to mimic 
a human dialogue partner when discussions turn to matters biographical. 

If the scope of the anticipated subsequent sentences is very narrow, one can
create a library of contexts and use a classifier to determine an
appropriate context choice for a given input sentence. This context can be
loaded and used to assign a meaning to the sentence with the help of
logical inference. To achieve this, the logical language to be used needs
to have the properties of completeness and compactness \citep{boolos:2007}. This means, 
however, that the expressiveness of both the sentence to be interpreted and the specification of 
contexts must be severely restricted --  thus they cannot  include, for example, 
intensionality\footnote{Predicates predicating over potentially non-existing entities.}, verb 
modality\footnote{For example, deontic assertions or wishful propositions} or second-order-
logic predicates\footnote{For example: `Mars is red. Red is a color.' (example from 
\citet{gamut:1991}). In the second sentence `is a color' predicates over the predicate `is red' 
from the first asentence.} -- thus marking one more dimension along which the machine will 
fall short of AGI\footnote{Deterministic workarounds are possible for the mentioned 
phenomena but they have nothing to do with AGI}.

What can be achieved in this fashion is illustrated in the field of
customer correspondence management, where there are repetitive customer
concerns that can be classified and for which pre-fabricated narrow
background contexts can be stored in the machine using first-order logic.
Customer texts can then be understood by relating them to this knowledge
base.\footnote{This is the approach described in section 3.2 of
\citet{landgrebeSmith:2019}.} However, it can be applied only in those
special sorts of situation where the relevant contexts can be foreseen and
documented in advance. 

When, in contrast, a machine has the task of engaging in dialogue with a human being, the
range of language production possibilities and of contexts and context combinations is as vast 
as the human
imagination. The human interlocutor can speak about anything he has
experienced, read about or can imagine, depending on his biography, his
current mood and intentions and their interaction with the situation he is
in. It is impossible to build a library of contexts that would prepare a
Turing machine for this kind of variation. In nearly all situations,
therefore, the machine will not have any context to load in order to assign
a meaning to the sentence, let alone to carry out routine tasks such as disambiguating 
\textit{personal pronoun anaphora} of the sort illustrated in a sentence such as:

\begin{enumerate}
\item[(5)] They caught a lot of fish in the stream, but one of \textit{them} died. 
\end{enumerate}

\subsubsection{Machine interpretation of suprasentential utterances} 

The space of possible contexts is all the more immense when we consider
multi-sentence (suprasentential) utterances. Here interpretation requires
the ability to identify and interpret complicated relationships between
sentences, including all the syntactic and semantic as well as explicit and
implicit sentence interdependencies of the sorts identified in
\ref{mssu}. In open text-understanding tasks\footnote{Closed tasks are
those in which a large proportion of the texts to be understood contain
repetitive patterns, such as customer or creditor correspondence or notices
of tax assessment.} it is impossible to foresee the possible sentence
relationships and to provide in advance knowledge of the sort that would
enable the machine to interpret them adequately.

Consider, to take a toy example, the tasks the machine would face in
interpreting the following sentences:

\begin{enumerate}
 \item[(6)] The salmon caught the smelt because it was quick.
 \item[(7)] But the otter caught it because it was slow.
\end{enumerate}

First, to understand that the explicit anaphora `it' refers to `salmon' in
both (6) and (7) --  even though two contradictory properties (`slow' and
`quick') are attributed to it -- and thus to understand the reason for the
adversative `but' in (7), the machine needs biological knowledge about the
species involved and about their respective hunting behaviours\footnote{The
interpretation of the second `it' as referring to the smelt is perhaps
still possible. Ambiguity is often simply not fully resolvable}. Given such
knowledge it can contextualise the two adjectives by tying them to
different parts of the total situation described in the sentence pair.
Already this is difficult -- but infinitely many such combinations with
much higher levels of difficulty are possible (for instance, consider this very text
which you, the reader, now have before you).

\paragraph{Machine non-interpretation} 
Another aspect that is difficult to model in a machine-compatible fashion is human conscious 
or unconscious non-interpretation of lexemes or phrases in dialogue, the phenomenon which 
Putnam calls `linguistic division of labour' (see Appendix \ref{nonInt}). How should a 
machine know whether it can afford to \textit{not} interpret a lexeme? 

\subsubsection{Machine interpretation of static non-lexematic material} \label{stt-nl}

As described above (see section \ref{criteria}), in a real conversation with a human, the machine has to interpret the 
entire structural material of an utterance, including the non-lexematic parts,
which means: facial expressions, gestures, body language, as well as sound
structures emanating from the interlocutor. Any of these can transform the
interpretation of the utterance conceived on the level of purely lexematic structures.

We will see that it is impossible for machines to detect such clues and to
combine them with lexematic material in a way that would make it possible
for them to achieve the sort of adequacy of interpretation that would be
required to lead an adequate conversation with a human -- for the reason that the variance
resulting from such combinations is effectively infinite, and each combination is a rare
event for which the needed training material could never be assembled in sufficient quantities. 
Furthermore, each combination of structural utterance material allows different
interpretations. To make a selection from them and to create a reply based
thereon that seems natural (non-stereotypical) to a human interlocutor
requires an array of capabilities and intentions rooted in experiences of manifold different 
sorts of contexts which the machine lacks.

\subsection{Modelling dialogue dynamics mathematically}\label{dynM}

In the previous section we have seen that it is very hard to make machines
utter and interpret single utterances. What happens in an entire, extended dialogue? As 
described in Appendix \ref{dyn}, the evolution of a dialogue
can be highly dynamic. The interlocutors switch roles as utterers and
interpreters as they take turns based on cues from their interlocutor in ordered or unordered
form (cutting each other short, interrupting, speaking at the same time).
While this is happening, their respective dialogue horizons are in constant
movement, and so are the intentions and speech acts based thereon. New
utterances interact with older ones, the dialogue creates its own context, see Appendix 
\ref{disDeix}.

From a mathematical perspective, a dialogue is a temporal process
in which each utterance produced is drawn from an extremely high
dimensional, multivariate distribution.\footnote{A multivariate
distribution is a distribution that can be modelled using the vector spaces
employed in stochastics. \citep{klenke:2013}} Each produced utterance can
relate to the utterances that preceded it in an erratic manner. In other
words: there is no way to formalise the relationship between the utterance
and what preceded it.\footnote{Falling in love at first sight is a classic
example of an event relating in an erratic manner to the events that
precede it. } Each utterance interpretation is drawn from a distribution of
similar complexity, and it too can relate also to the utterance that preceded it
in an erratic manner.

To see the sorts of problems that can arise, consider a dialogue between
Mary and Jack spanning several rounds of role-switching. Mary makes an
utterance at round 7, which requires Jack to take into account an utterance
from round 3. Based on this, Jack associates with Mary's utterance an
experience from his own past, of which Mary knew nothing, and he provides an
answer relating to this experience. This utterance from Jack is for
Mary quite unexcepted (erratic) given her utterance in round 7. But, given his inner
experience, it is perfectly coherent from the perspective of Jack. Phenomena such as this imply 
that there is no way to formalize the relationship between an interpretation of an utterance and 
this utterance itself. Such phenomena break the Markov assumption employed by the relevant 
temporal process models.

\subsubsection{Modelling dialogues as temporal processes}

To understand the dialogue as a temporal process, four types of events need
to be distinguished:
 \begin{enumerate}
\item initial utterance production, followed by 
\item initial utterance interpretation, followed by 
\item dialogue-dependent responding utterance production, followed by
\item dialogue-dependent utterance interpretation. 
\end{enumerate}
These are linked via relations of dependence. In the prototypical case,
pairs of events of types 3. and 4. are repeated until the dialogue
concludes.

The distribution from which each of these events is drawn varies massively
with the passage of time, as ever new utterances are generated and
interpreted.

Both utterer and interpreter have a huge number of choices to make when
generating and interpreting meaning, and because these choices depend on
the diverse dialogue contexts (including the dialogue itself) and on their
respective horizons, as well as on the biographies, personalities,
capabilities, intentions, (and so forth), of the participants themselves, it follows that each 
utterance and each interpretation thereof is \textit{erratic}. Such an event, like the nuclear 
fission event occurring in radioactive decay, is unrelated to the events that precede it, it is  
purely random, which means: it cannot be modelled as depending on what occurs in the 
immediately preceding dialogue step.

To make matters worse, we still have not taken into account the fact that
most human dialogues deviate from the turn-taking prototype, and 
it is not conceivable that we could create a mathematical model that would
enable the computation of the appropriate interpretation of interrupted
statements, or of statements made by people who are talking over each
other, see Appendix \ref{ellipse}.

 Or consider the problem of computing the appropriate length of a pause in a conversation (or, 
equivalently, of inferring from context the reason why your dialogue partner is not responding 
in a timely manner to what you have just said). Appropriate pause length may depend on 
context (remembrance dinner, cocktail party), on emotional loading of the situation, on 
knowledge of the other person's social
standing or dialogue history, or on what the other person is doing (perhaps looking at his 
phone) when the conversation pauses. Pauses are context modifiers which influence or are 
important ingredients of the overall dialogue interpretation. They often contain subtle non-
verbal cues, for example, the fiddling of the interlocutor with a small object indicating
irritation or nervousness. The machine must somehow assess all of these factors to determine 
how it should react to the pause -- which might signify that for the interlocutor the dialogue is 
at an end, or that he is inviting a break in the expended role-change cycle, or that he is 
engaging in a battle of wills. To be done properly this assessment requires both (1) a human 
background of life-long experience, and (2) an intention to achieve something by reacting to 
the pause in a certain manner, for example: to heal the breach, to win the battle of wills, and so 
forth. Machines lack both.

\subsubsection{Mathematical models of temporal processes}
On the other side of the ledger, the range of options for mathematical treatment of dialogue is 
strictly limited. In fact there are only two types of explicit mathematical methods available to 
model temporal
processes: \textit{differential equations} and \textit{stochastic process
models}. Given that there are no other methods available,\footnote{That is,
nothing else exists in the currently available body of knowledge of mathematics and theoretical
physics.} any Turing machine able to model such processes would have to
draw from these alternatives or from some combination thereof.

\paragraph{Differential equation models}
Such models can be used to provide adequate
representations of the changes in related variables when their
relationships follow deterministic patterns of the sort that can be
observed in the physical realm (for example radioactive decay over time).
By ``adequate model'', we mean a scientific model that is able (in
ascending order of scientific utility) to 1. describe, 2. explain or 3.
predict phenomena and their relationships \citep{weber:1988}, the latter with different 
degrees of accuracy.

Description is the minimum requirement for any scientific model, but the other
two properties are also needed to make the model useful. Differential
equations can be used, for example, to make predications regarding changes
such as are involved in the distribution of heat from a source in space, something that
is modelled using the so-called heat equation, which is a partial differential equation
first developed by Fourier in 1822.  But such models can only deal with a
small number of variables and their interrelations, and they are of the sort that can be verified 
using physical experiments. 

To conduct an adequate conversation, plausible utterances have to be produced by the
machine. Mathematically speaking, an utterance produced by the machine -- no matter what 
algorithm is used -- is a model-based prediction conditioned on the previous
utterances in the dialogue and on the context. The machine predicts what the next move in the 
dialogue should be, just as a Go-playing machine predicts its own next move conditioned on 
the opponent's last move and the overall situation on the board). 

We note that the situation is different in the case of a human dialogue partner, where humans 
do not necessarily need to predict what their own next move in the dialogue will be because 
they themselves are deciding that next move on the basis of their intentions - however, their 
response to a given utterance corresponds to a prediction computed by a machine based on a 
given utterance. But because the machine has no intentions and life-experience, it will not be 
able to compute an adequate response, and even less to predict the reactions of the interlocutor 
in the way that humans (in many cases) can. The machine will therefore have a massively 
shallower basis for selecting an appropriate utterance.

At the same time, however, the machine-predictions would need to be highly accurate (where 
accuracy, for stochastic models is measured by the percentage of predictions that match 
human expectations). In the case at issue here, this would mean a high degree of utterance 
salience. Further, their accuracy in this sense has to be maintained over the entire course of the 
dialogue, otherwise the first criterion of the test will fail because the human will have to make a 
conscious effort of the sort that is associated with the need to interact with a machine.

It is a problem therefore, that differential equations cannot even
provide descriptions, much less explanations or predictions, of the changes
involved as concerns social processes in particular and
biological phenomena in general. This follows already from the fact that the number of 
variables involved in such phenomena is too large, and their interdependences too
complex, to make such modelling possible. Evidence that proposed models do
not work in these sorts of contexts is provided by the fact that they are repeatedly
falsified by empirical observations. Where differential
equations are used successfully in biology this is because the number of
variables has been limited artificially, for example when organism population growth
is modelled under simplified laboratory conditions.

The application of differential equation-based models to the problem of
dialogue production and interpretation is for the same reason impossible.
There are far too many variables, and we cannot even begin to formulate
equations that would describe their relationships. The reason for this is
that, although all the parts of the brain function in accordance with the
laws of nature, the system behaviour is hypercomplex \citep{thurner:2018} in each of its 
behavioural patterns, \footnote{A hypercomplex system obeys deterministic laws but cannot 
be mathematically modelled due to overcomplexity.} which include all the phenomena of 
language production addressed in the in Appendix An erratic event, on the other hand, cannot 
be modelled using differential equations \citep{schuster:2005}. We cannot even describe it in 
these terms, much less obtain predictions.

\paragraph{Stochastic process models} Stochastic process models can represent the behaviour 
of a one- or multi-dimensional \textit{random} stochastic process $X$, but only if
\begin{enumerate}
\item the random event, and thus the associated random variable (in what follows: r.v.) $X_t$, 
has a distribution over time belonging to the exponential family,\footnote{Often this is the 
Gaussian distribution, i.e. $X_t \sim \mathcal{N}(\mu,\,\sigma^{2})\,.$}
\item the process has additional properties that allow mathematical modelling (specifically, it 
must have independent and stationary increments, as further specified below).
\end{enumerate}
The most expressive family of stochastic models, and thus the models that
have had the widest usage in describing phenomena based on human
interactions, are the Wiener process models (also referred to under the
heading ``Brownian motion''), which have been used
extensively (indeed, too extensively, as we shall see) to model financial
market processes such as movements in stock or derivative prices
\citep{jeanblanc:2009}. Such prices are an expression of the aggregated
intentions of very many market participants. The models make strong
mathematical assumptions, for example that a price change process $X$ is a
case of Brownian motion, or in other words that it satisfies the following
conditions:
\begin{enumerate}
\item it has independent r.v. increments: for any pair of time
points\newline $(s,t)$, $X_{t+s} - X_s
\perp \!\!\! \perp \mathcal{F}^X_s $, where $\mathcal{F}^X_s$ models the time before $t$,
\item it is stationary:  $\forall s>0: (X_{t+s}) = (X_t), t \geq 0$, and
\item for any time point  $t > 0$, $X_t \sim \mathcal{N}(0,t)$.
\end{enumerate}

Condition 1. expresses the fact that each increment of the r.v. is
independent of what happened in the past; condition 2. that the
unconditional joint probability distribution\footnote{`Unconditional' means that the 
distribution involves no dependence on any particular
starting value.} of the process does not
change when shifted in time; condition 3. expresses the fact that the r.v.
is distributed according to the Gaussian distribution.

Unfortunately, processes satisfying these conditions are nowhere to be
found in actual markets. This is, again, because the preferences and
intentions of human beings are erratic (in part because they depend on real
world events, for example geopolitical events, which are also erratic).
This is why, whenever collective decisions are off-trend, financial
stochastic process models fail \citep{mccauley:2009}.

Dialogues, too, as we have seen, are multivariate processes, with the r.v.
-- utterances and interpretations -- drawn from immense, typically unknown, and
in any case not modellable, multivariate distributions. Neither utterances
nor interpretations are distributed according to a multivariate Gaussian
distribution, since they are non-stationary and non-independent. And, to
make matters even worse, interpretations are not directly
observable (see Appendix \ref{impl}).

The Brownian motion model is therefore not applicable, as none of its three
conditions is satisfied. 

\subparagraph{Hidden Markov Models}

A hidden Markov model (HMM) is a stochastic model which models a process as a Markov 
chain where successive observable events are generated by transitions between unobservable 
states.

If a dialogue would meet (1) the cardinal assumption of an HMM, namely satisfaction of the 
Markov property, together with (2) the assumption that transition probabilities remain 
constant over time, then it could in theory be used to model dialogue utterances as emanations 
from those unobservable mental events that lead to the utterance generation and 
interpretation. Unfortunately, HMMs cannot be used to model dialogues, since dialogues 
violate both assumptions.

\paragraph{Stochastic differential equation models}
Differential equations can be extended to model temporal processes
subjected to stochastic effects (noise), for example to model molecular
dynamics. Again, however, even stochastically modified differential
equations would still not be applicable to the problem of dialogue process
modelling, since this would require that the assumptions needed for the
applicability of \textit{both} differential equations \textit{and}
stochastic process models would need to hold simultaneously for processes of
language use. In fact, however, both of these sets of assumptions fail.

\paragraph{Deep Neural Network (dNN) models} \label{stochPM}
These are a subclass of stochastic models that
in recent years have sparked considerable enthusiasm,
triggered above all by:

\begin{enumerate}
\item the successes achieved since 2014 in improving automated translation through use of 
dNNs, 
\item the popularisation by \citet{goodfellow:2014} of generative adversarial networks 
(GANs),\footnote{Invented and first described by \citet{schmidhuber:1990}.} and 
\item the invention of reinforcement learning, which brought the capability to outperform 
human beings, for example in the game of Go \citep{silver:2016}. 
\end{enumerate}

dNNs were accordingly tested early on in the domain of process modelling.
They differ from classical mathematical models, which are explicitly designed
in a conscious mathematical effort, for example when observing process
data and figuring out an equation to desribe them. In contrast to this,
dNN-models are created automatically by an optimisation algorithm which is merely
\textit{constrained} by humans. As is also the case with traditional
multivariate regression models, which have been used routinely since the 1970s
\citep{hastie:2008}, the optimisation algorithms can create new models that humans would
not be able to construct when modelling explicitly. These
dNN-generated-models are \textit{automated}. The
equations they consist of are not created by human effort, but rather by the
optimisation algorithm working under constraints (for example the loss
function and the hyperparamters of the dNN). How the resulting equations (which can be
inspected) solve the machine learning problem at hand often cannot be understood by
humans -- hence the explainability problem of AI \citep{goebel:2018}. However, this ability to auto-compute
models does not mean that machines develop intentions -- the
equations are just functionals or operators relating an input vector to a
certain output -- in other words, they are nothing but a special case of regression models.

We review the potential capability of three seemingly promising dNN-methods
to model human dialogues, before looking at the empirical evidence yielded
by experiments in dialogue emulation.

\paragraph{Deep recurrent neural network (rNN) models} \label{rNN}
Deep recurrent neural networks are dNNs in which the connections
between the nodes of the dNN graph allow the modelling of
temporal sequences. They are often called sequence-to-sequence-dNNs, because
they can be used to create one sequence from another (for example, a
translation from an input sentence). Often long-term-short-term-memory
(LSTM)  \citep{schmidhuber:1997} and its numerous extensions are used in
practical AI-applications of this sort, including Google Translate.
Because classical stochastic process models are not able to model
multivariate processes, the ability of rNNs to model temporal processes of this sort has
been investigated in recent years as a potential saving alternative
\citep{dasgupta:2016, neil:2016, lai:2018}. The results have performed well
for certain sorts of tasks, for example modelling road traffic
occupancy, solar power production, or electricity consumption over time
\citep{lai:2018}. As the latter reports, they have outperformed classical
stochastic process models in certain tasks, especially when two
processes with different patterns are overlaid in a series of observations.

We can infer from these examples several reasons why rNNs work well on such
numerical time-series data:

\begin{enumerate}
\item data of these sorts approximately fulfil the assumptions needed for stochastic process 
modelling in general (of which dNNs, and thus rNNs, are a special case), 
\item the data are \textit{repetitive} and huge historical datasets are available for training 
purposes, 
\item the dimensionality and the variance of the data is low,\footnote{Exchange rates, another 
example modelled using dNNs, form a special case. This is because outcome dimensionality 
and variance are here relatively low in the short term. Unfortunately, mid-term outcomes are 
erratic, and thus the models work less successfully. \citep{lai:2018}.} 
\item dNN architectures can be used to model temporal pattern overlays of the sort observed 
for example in traffic occupancy, which has both a circadian and a workday vs. weekend 
rhythm.  
\end{enumerate}

Human dialogues, however,

\begin{enumerate}
\item [i.] are not repetitive, but erratic;
\item [ii.] do not fulfil the central assumptions presupposed by temporal
process models which must also be satisfied for rNN to succeed in this modelling
task;
\item [iii.] are of extremely high dimensionality; and 
\item [iv.] manifest variance that is as large as the sum of the results of all human activities 
since the emergence of our species. 
\end{enumerate}

Moreover, because the interpretations involved at each stage of the
dialogue are implicit (see \ref{impl}), we can never use the interpretation step in human 
dialogue as a source of training data.

This will mean that there can never be training data to
cover the dependencies that hold between successive utterances occurring
over time, since interpretations are an essential link in the dependency chain that binds 
one utterance to its predecessor.

Note again, however, that all of this holds only for dialogues \textit{in general},
the mastery of which is a criterion of AGI. As we shall see, for
very stereotypical dialogues, for example the telephone scheduling of a
haircut or the reservation of a hotel room, there could eventually be
sufficient training data for a dNN-based approach to be of value.

\subparagraph{Generative adversarial network (GAN) models} \label{GANs}
\hspace{-11pt} work using two networks, one discriminative, the other
generative \citep{goodfellow:2014}. The former is trained
to discriminate classes of input data using annotated training material,
often pictures tagged by human beings (for example pictures in which humans
can be distinguished from other items represented). The generative network
is then tasked to create new samples of one desired class (for example pictures
of humans, \citep{karras:2018}, which it can indeed do). The two networks
are then chained together
by having the samples yielded by the generative network passed on to the
discriminative network for classification. Finally, the system is optimised to
minimise the rate at which samples are generated that are not classified by
the discriminative network as belonging to the desired class. This approach
works very well with pictures, because the discriminative net can be
pretrained with adequate training material (data tagged by humans).

Again, however, GANs are not applicable to language in the form that we encounter it in 
general spoken dialogue. For to build an
utterance-generating GAN that creates meaningful output one would need to
pretrain a discriminative net that can distinguish meaningful from
non-meaningful utterances. The problem is that, because the meaningfulness
of an utterance depends on its context and interpretation, there is again no
conceivable way in which a sufficient body of training material could be
assembled to cover the practically infinite variance of human utterances.
It is therefore not possible to create a discriminative net that can be
used to build a meaningful-utterance-producing GAN.

\paragraph{Models based on reinforcement learning} \label{rl}

In reinforcement learning \citep{sutton:2018}, a reward (score) is assigned when a certain step
in a repeatable type of finite process is realized by the machine.
`Finite', here means that the process ends after a series of steps that is
not too long, such as a game of \textsc{Go} or a first-person shooter game
in which killing sequences are repeated. In \textsc{Go}, for example, a
trained algorithm is used to assign a score after each action the machine
performs in each game. The machine obtains one point for each of the opponent's stones it 
captures and one point for each grid intersection of territory
it occupies. The trained algorithm is optimised to maximise the total score
obtained over the entire game. This is done by having the computer play the
game millions to billions of times in different situations, so that optimal paths for
these situations can be found and stored in the model.

Crucial, for such optimization to be possible, is that the scores for every
move can be assigned automatically by the machine. Machine learning of this
sort can thus be used only in those situations in which the results of
machine decisions \textit{can be scored through further machine decisions}. This is
primarily in games, but the method can be extended, for example, to debris
cleaning, where what is scored is the number of units of debris
removed. In such narrowly defined situations, machines can find strategies that outperform 
human behaviour \citep{jaderberg:2018}. Lastly,
reverse reinforcement learning
\citep{arora:2018}, a technique to automatically learn an adequate reward
score from observed situations, does not help in the dialogue case, because
there is here no adequate set of observed situations since, again, too few patterns are
repeated sufficiently many times. 

Reinforcement learning cannot, therefore, be applied to the engineering of
convincing dialogue systems. There is here nothing to which
the needed sorts of scores can be assigned. (There is no \textit{winning},
as we might say; or at least no winning of the sort that can be generally,
and repeatedly, and consistently, and automatically scored.) We note also
that the truly impressive successes of reinforcement learning do not
provide evidence that AGI is about to be achieved. This is because
the scope of applicability of such algorithms is narrowly limited to
situations in which automatic scoring \textit{is} possible. It is also because the
meta-parameter for the algorithms which compute the optimisation, including how its scores 
(and many other parameters) are to be defined, needs
to be set in each case by engineers.

\subsubsection{Current state-of-the-art in dialogue systems: A review of what has been 
achieved thus far} 
Even given all of the above, dialogue emulation is an area of considerable
activity in AI circles. The resultant dialogue systems -- also called
`agents' (or in some circles `chatbots') -- are designed and built to
fulfil three tasks (citing \citep[6]{gao:2018}):

\begin{enumerate}
 \item Question Answering -- ``the agent needs to provide concise, direct answers to user 
queries based on rich knowledge drawn from various data sources'' 
 \item Task Completion -- ``the agent needs to accomplish user tasks ranging from restaurant 
reservation to meeting scheduling $\dots$ and business trip planning''
 \item Social Chat -- ``the agent needs to converse seamlessly and appropriately with users -- 
it is performance along this dimension that defines the quality of being human -- and provide 
useful recommendations''\footnote{To this belongs the ability to adjust the dialogue horizon to 
the dialogue partner, for example to adjust their respective intentions.}.
\end{enumerate}

In our view at least, the third task could only be performed by a machine
with AGI. Indeed, it would precisely be one of the purposes of AGI to
perform tasks of this sort. 

\paragraph{Question Answering and Task Completion} 

Question Answering and Task Completion are areas in which dialogue systems
are already of considerable commercial value, mainly because customers with
relatively homogeneous cultural backgrounds can be motivated to reduce
their utterance variance --  for example by articulating clearly and using
sentences from a pre-determined repertoire -- if by interacting with a bot
they can quickly obtain answers to questions or resolution of boring tasks.
In the medium term, technologies of these sorts will enable systems which
can satisfy a double-digit percentage of customer requests.

However, Question Answering and Task Completion clearly have nothing to do
with conducting conversations in a way that would be indicative of AGI. Fulfilment of each is 
(thus far) something that is achieved simply by using appropriately configured software tools, 
which every user identifies as such immediately on first engagement.

\paragraph{Social Chat} \label{chat}

What, then, about social chat (also called `neural chitchat') applications?
Here, research is currently focused on two approaches:

\begin{itemize}
 \item  supervised learning with core technology end-to-end
 sequence-to-sequence deep networks using LSTM (section
 \ref{rNN}) with several extensions and variations, including use of
 GANs,\footnote{Discussed in \ref{GANs} above, and in \citep[53-56]{gao:2018}} and 
\item reinforcement learning used to train conversational choice-patterns over time (the 
optimal path of machine utterances during a dialogue).\footnote{Discussed in \ref{rl} above, 
and \citep[59-61]{gao:2018}.} 
\end{itemize}

Strong claims are made on behalf of such approaches, for example in
\citet{zhou:2018}, which describes Microsoft's XiaoIce system -- ``XiaoIce''
is Chinese for ``little Bing'' -- said to be ``the most popular
social chatbot in the world''. XiaoIce was ``designed as an AI companion
with an emotional connection to satisfy the human need for communication,
affection, and social belonging.''  The paper claims that XiaoIce
``dynamically recognizes human feelings and states, understands user
intents, and responds to user needs throughout long conversations.'' Since
its release in 2014, XiaoIce has, we are told, ``communicated with over 660
million users and succeeded in establishing long-term relationships with
many of them.''

Like other ``neural'' chitchat applications, however, XiaoIce displays two
major flaws, either of which will cause any interlocutor to realise
immediately that they are not dealing with a human being and which will
prevent any sane user from ``establishing a long-term relationship'' with
the algorithm.

First, such applications create repetitive, generic, deflective and bland
responses, such as ``I don't know'' or ``I'm OK''.  This is because the
training corpora they are parametrised from contain many such answers, and
so the likelihood that such an answer might somehow fit is rated by the
system as high. Several attempts have been made to improve answer quality
in this respect, but the utterances produced by the algorithms are still
very poor. The reason is that the algorithms merely \textit{mimic} existing
input-utterance-to-output-utterance sequences without \textit{interpreting}
the specific (context-dependent) input utterance the system is reacting to.

Each input is treated, in fact, as if it were the input to a machine
translation engine of the sort which merely reproduces sentence pairs from
existing training sets. The difference is that here the training sets
consist of pairs of sentences succeeding each other in one or other of the
dialogues stored in a large dialogue corpus. The result is that, with the
exception of a small subset of the structural elements, none of the sources
of human discourse variance listed in section \ref{vars} are taken into
account in generating output utterances. Again, no attempt is made to
\textit{interpret} utterance inputs. Rather, the machine in responding
simply tries to copy those utterances in the training set which immediately
follow syntactically and morphologically similar input symbol sequences.
This means that utterances are decoupled from context, and so responses
appear ungrounded. Attempts to improve matters using what are
called ``Grounded Conversation Models'' \citep[section~5.3]{gao:2018}-- which try to include 
background- or context-specific knowledge -- have not solved the problem. The failure to 
model the variance of the utterance sources persists.

Second, these sorts of applications create ever more incoherent utterances
over time. This is first of all because they cannot keep track of the
dialogue as its own context (Appendix \ref{dc}), and secondly because the datasets
they are trained from are actually \textit{models of inconsistency} due to
the fact that they are created as mere collections of fragments drawn from
large numbers of different dialogues. Attempts to alleviate the problem
using ``speaker'' embeddings or ``persona''-based response-generation
models have improved the situation slightly \citep{ghazvininejad:2017}, but
they do not come close to ensuring realistic, convincing conversations.

Given that machines of the mentioned sorts can neither \textit{interpret}
utterances by taking into account the sources of variance, nor
\textit{produce} utterances on the basis of such interpretations together
with associated (for example biographical) knowledge, the approach cannot
be seen as promising when it comes to conducting convincing conversations.

\paragraph{Reinforcement learning in neural chitchat}
The basic problem of reinforcement learning (RL) for social dialogue is
that it is impossible to define a meaningful reward. XiaoIce itself uses CPS
(conversation turns per session, \citep{zhou:2018}), a measure that
maximises the duration of a conversation. We doubt, however, that a human
would be impressed by dialogue behaviour generated to optimise a measure of this sort.

\citet{li:2016} used a more sophisticated reward system by training an
RL-algorithm using dNN-generated synthetic utterances (because using real
human utterances would be prohibitively expensive) together with a
tripartite reward function rewarding
\begin{enumerate}
\item non-dull responses (using as benchmark a static list of dull phrases such as ``I don't 
know'')
\item  non-identical machine utterances, and 
\item  Markov-like short-term consistency. 
\end{enumerate}
The results are appalling, and one wonders why this type of research is
being conducted at all, given that -- as a result of its use of synthetic data -- it
violates the basic principles of experimental design as concerns adequacy
of measurement setup for observation of interest.

\paragraph{Multi-purpose dNN language models} \label{mp}

Recently, \citet{radford:2018} created multitask dNN-language models from
large corpora by formulating the learning task as the ability to predict a
language symbol -- for example a single word -- based on the symbols
preceding it. These models were trained using an unsupervised approach, but with the 
possibility to condition the model on certain task types \citep{mccann:2017}.\footnote{In
the type of unsupervised learning described there, the algorithm learns
models of probability distributions for symbol sequences from unlabelled
input data. These models reflect symbol sequence distributions. Once
created, they can be used to predict symbol sequences given conditioned
input.} The model that results (dubbed ``GTP-2'') is then conditioned with
problem-specific input data to produce model-based predictions to solve NLP
benchmarks (``zero shot predictions''). For some basic tasks amenable to
sequence-modelling (including translation and text gap filling) the
performance is good. For question answering, however, which is the only
dialogue-related task that was tested, only 4.1\% of
questions\footnote{Typical example: ``Largest state in the US by land
mass?''} were answered correctly.

\subsubsection{Problem-specific AI: Turing machines enriched by prior knowledge}

Looking at the main problem of social chatbots, namely their inability to
interpret utterances and to react to them with context-adequate, biography-
and knowledge-grounded responses, one could indeed imagine endowing an
algorithm with systematic prior knowledge of the sort required for
conversations. The system presented in the Appendix to \citet{landgrebeSmith:2019}
incorporates prior knowledge in this way, focusing on knowledge of the sort
needed to complete tasks such as simple letter and email answering or
repair bill validation. It uses this prior knowledge and logical inference
in combination with machine learning to explicitly interpret texts on the
basis of their business context and to create adequate interpretation-based
responses. However, it works only because it has strong, in-built
restrictions.

\begin{itemize}
\item The range of linguistic inputs it has to deal with is very narrow
(for example car glass damage repair bills or customer change of address
requests), thereby avoiding the problem of complex and nested or
self-referential contexts (section \ref{context}).
\item It is not a conversation system and does not have to model a
stochastic process, because it reacts always to just one language input,
thereby avoiding the problem of temporal dynamics (see Appendix \ref{dyn})
-- a problem that is not amenable to mathematical modelling (see section
\ref{dynM}).

\end{itemize}

Such a system would fail in dialogues, and this would be so even if it was
stuffed to the gills with (for example) biographical knowledge of the
dialogue participants. This is because it could not cope with either the
complex dialogue contexts or the dialogue dynamics. These phenomena
aggravate the difficulties in dealing with language economy, dialogue
structure and modality, because the contexts and the dynamics create a huge
range of interpretation possibilities on all such levels. The resultant
infinite variance makes it impossible to provide the machine with sufficient
knowledge to derive meaningful responses.

\section{Conclusions}
How, then, do \textit{humans} conduct convincing conversations? Answer: by using language, 
as humans do. Language is a unique human ability that evolved over millions of years of 
evolutionary selection pressure. Using language gives us the ability to realize our intentions, for 
instance by generating initial utterances (engaging in dialogue as a means of expressing ideas 
or desires) and by dynamically interpreting an interlocutor's utterances. This then allows us to 
react adequately, either with further utterances or with corresponding actions. Subconsciously 
or consciously, a human interlocutor is thereby able to sense the purposes of a fellow human
being with whom he interacts because this is a survival-critical ability. As Gehlen points out, 
using language effectively in a given situation requires the dynamic exploitation of our past 
experiences, both inner and outer, as these have become engrained within our neural 
substrate. 

Using language in the way that humans use language cannot be conceived without a body of 
the sort that has grown up in a world of sensory experiences and practical agency. Machines 
lack this type of experience and they lack any framework of intentions that could shape the way 
in which they interpret or generate utterances. 

Ultimately, most human interlocutors will notice that a machine has no intentions because of 
its inability to react properly to a dynamic conversation. An analogous lack of intentions and 
purpose can be experienced when speaking to long-term schizophrenics with acquired autistic 
syndrome. Their reaction patterns are immediately perceived as non-normal because their 
ability to interpret and create utterances has deteriorated \citep{bleuler:1983}. Machines 
perform much worse than do such patients, and therefore most interlocutors will rapidly sense 
their ``non-normality''. 

The AI community has so far failed to come to grips with the physical, bodily side of human 
language production and interpretation and the infinite landscape of variance in dialogue 
utterances which it brings in its wake. Could they take these factors into account with new 
system designs? We have argued that there is no way to mathematically model the human use 
of language. Certainly novel approaches such as adversarial dNN and reinforcement learning 
paradigms have enabled the creation by the machine of novel algorithms, which are notable 
exceptions to Ada Lovelace's proposition that a Turing machine cannot learn anything new. 
But as we have seen, they will not learn to speak as humans do, because what they can generate 
is by far too restricted to emulate human language ability. This would still be so even if we 
could make available the huge quantities of data -- orders of magnitude greater than the 
datasets used to train Google Translate -- that such a model would need if a machine was to be 
trained to implement it.

Until the time is reached where a type of mathematical model is proposed that would be in a 
position to represent the dynamic properties of human dialogue, we believe that the idea that 
the ability to use language properly will somehow emerge spontaneously in the machine when 
storage and computing power reach a certain threshold will remain a product of magical 
thinking. Our understanding of the human brain, and of the evolution of the human physical 
substrate, and of how this physical substrate is shaped by what the individual learns from its 
surrounding culture, will, to be sure, increase tremendously over the decades and centuries to 
come. But this physical substrate will remain a complex system in the sense of 
\citep{thurner:2018}, and so it will remain subject to the same fundamental obstacles to  
mathematical modelling that we have already described. 

We do not, however, wish to imply that only something with our kind of DNA, neurons, and so 
forth, could conduct convincing conversations. We think that any entity with real intentions 
and the ability to undergo auto-modifications analogous to those inherited changes of genotype 
which have affected modern human beings and their ancestors over some 3 million years 
could evolve to conduct convincing conversations given enough time and environmental 
pressure.

\appendix

\section{Appendix: Variance and context in human dialogue}\label{vars}

\subsection{Levels of language production and interpretation}

To document the enormous potential for variation in human dialogue
interactions, we describe in detail the different levels on which the
context and structure of a dialogue and the form of its dynamic
interaction processes are determined. 

Loosely following \citet{verschueren:1999}, we distinguish five levels of
\textit{language production and interpretation}, namely:
\begin{enumerate}
      \item context, 
      \item language economics (deixis and implicit meaning),
	  \item dialogue structure (words, sentences, gestures, $\dots$)\footnote{We follow 
Verschueren (\textit{op. cit.}) in using the term
``structure'' to designate what might otherwise be referred to as
``content'' or ``material''. Utterance structure can be both verbal and
non-verbal (for example when it involves use of gestures).},
      \item force/modality,
      \item dialogue dynamics. 
\end{enumerate}

When humans engage in conversation all of these levels interact. Their
separate treatment here is necessary merely in order to enable systematic
description; in reality they can never be properly spliced apart.

\subsection{Types of dialogue context} \label{context} 
The \textit{dialogue context} is a ``setting'', where this term is to be
understood in a broad sense to embrace, for instance: one's place at the
dinner table, one's place in society, a geographical place, the time of day
at which a dialogue occurs, and many more \citep{barker:1968}. In each case
the context is determined by an interplay between the wider environment and
the identities of the parties involved, including their mental attitudes, capabilities, and
intentions.

\subsubsection{The dialogue horizon}
Dialogue contexts are marked not by sharp boundaries but by what
is called a ``horizon'' of possibilities, for example the possibility that
our dialogue partner might be lying, or intending to report our
conversation to his superiors. The horizon of a spatial context might include the possibility of 
leaving through the back door;
the horizon of a temporal context that one's husband may return at any
moment. For each interlocutor, the dialogue context is thus in some ways
analogous to the visual field of an individual subject: now more things, now
fewer things fall within its compass. And the things which do fall within
its compass do so in a way that encompasses a penumbra of
possibilities.\footnote{Compare Husserl: ``The world is pregiven to us, the
waking, always somehow practically interested subjects, not occasionally
but always and necessarily as universal field of all actual and possible
practice, as horizon.'' In our natural, normal life ``we move in a current
of ever new experiences, judgments, valuations, decisions'', in each of
which consciousness ``is directed towards objects in its surrounding
world'' surrounded by a horizon of fluently moving potentialities
\citep[142,149]{husserl:1989}.} Consider for example how facial
expressions become apparent as we move closer to persons in our visual
field, and how these facial expressions themselves bring to light new
potentialities for greeting and embracing. 

In each dialogue, each dialogue participant will have at any given stage a 
\textit{dialogue horizon}, which results from the combined effects of all his salient dialogue 
contexts at that stage. 

This dialogue horizon encompasses all possibilities that fall within the scope of what is relevant 
to the interlocutor, as determined not only by his identity, and by his intentions of the moment, 
but also by the social and cultural setting of the dialogue and by other contextual factors. The 
way each interlocutor shifts his intentions alters his dialogue horizon, which in turn 
determines how he perceives new utterance material. This then has a dynamic effect on new 
intentions, which further shape his interpretation and the way new speech acts are formed and 
new contexts for interpretation are created.

\subsubsection{Social, cultural and environmental contexts} 

\paragraph{Social context} \hspace{-11pt} is the social setting of the
conversation \citep{hanks:1996}, for example
the context of a family outing, of two strangers bumping into each other on
a railway platform, of a teacher berating a failing student, of a session
in parliament. As the latter cases make clear, a social context may include
institutional elements, and in such cases we can refer also to an
institutional context. The social context exists in virtue of the fact that
the participants in the conversation have formally or informally defined
\textit{roles} in virtue of which they are subject to certain norms. The
social and institutional rewards and sanctions associated with these norms
then form part of the dialogue horizon. They influence not only what the
dialogue partners say (and what they do not say), but also the ways they
speak and act.

\paragraph{Cultural context} \label{cultc} \hspace{-11pt} is a special sub-type of social
context. It is the setting created by those socialisation patterns which
come into play where the participants in a dialogue draw on a common
cultural background passed on from one generation to the next. The cultural
context is thus determined by those habits, norms and values which result
from similar types of upbringing, education, and so forth. 

The social context of a conversation constrains in each case the space of
permissible utterances. A relatively open space is obtained where social peers
speak in private; a much narrow space arises when institutional or social inferiors and
superiors speak in an institutional setting, for example judge and
defendant in a criminal case. (We note that even here both parties will
sometimes step outside the institutionally accepted norms. As in every
other type of dialogue, the possibility that a participant forms the
desire, for example, to shock or bamboozle his interlocutor can never be
ruled out.) On the other hand, if dialogue partners do not share any
cultural context or tradition, and do not know about each other's social
roles, then they will likely choose a very general communication context
that is appropriate simply for an encounter between fellow humans. Even here, however, there 
is no simple recipe to determine what communication context will arise. This may turn on the 
fact that the interlocutors belong to the same age cohort, or that they are waiting on the same 
railway platform for the same train. Ad hoc features of this sort can affect all context selection  
in a way that cannot be predicted in advance, for example by some algorithm.

\paragraph{Contextual constraints on language use} There is a variety of
social contexts which constrain our dispositions and choices when producing
language, and conversation participants may engage one or more of these
within a single conversation. Each determines a particular variety (a
``code'' or ``register'') of the language used in the conversation. A
\textit{sociolect} is an expression of the constrained dispositions and
choices of those language users who share a social background resulting
from a shared pattern of socialisation. Age cohorts also have sociolects,
as do members of specific criminal gangs. 

A \textit{dialect} is a sociolect of those language users who share a social
background that is regionally determined. A \textit{grapholect} is a
written language as standardized for example in a dictionary. A
\textit{cognolect} reflects the constraints imposed on an utterer by her
intellectual abilities and education level, which may include a common
professional or disciplinary socialization in, for example, architecture or
rap music.

\subsubsection{Spatial and temporal context}
Spatial context is the \textit{site} of the dialogue, formed by the
physical place (the park bench, spaceship, bus, hospital, pub, bed, and so
on) in which it takes place. Temporal context is the time (dusk, Christmas,
tea break) in which the dialogue takes place. Both temporal and spatial
context can include (at several levels) other spaces and times nested
within them, for instance when a dialogue relating to the food on the dinner table
suddenly switches its context as the diners become aware that someone is
banging hard on the front door, or when a
dialogue happens at one time but the interlocutors speak about other times
and about their temporal order. Consider a conversation between a police
officer and the various parties, including witnesses, involved in a car
accident. Consider such a conversation where, among the various parties,
there are some who speak different languages.

Both spatial and temporal context are determined in part by the
\textit{communication channel} used in the dialogue. This can be local, in case of face-to-face 
communication, or remote. It can be spoken versus
written, and direct versus asynchronous, with different degrees of delay
(such as chat -- text message -- email -- letter).  Skype combines
verbal, visual and textual (chat) elements, and both of the latter can be
enhanced in turn with emojis. Again, there are different sorts of rules and
norms associated with different sorts of channel, and different channels
are more or less adequate or appropriate to different sorts of
communication. A text message channel may be adequate for announcing one's
arrival time; not however for expressing condolence on the occasion of
someone's death \citep{westmyer:1998}. 

\paragraph{Environmental context} \hspace{-11pt} is the setting formed by that part of the 
world in which the conversation takes place. It is a combination of spatial and
social context, and thus includes both physical and social constraints. It
is made up of what Barker calls ``ecological units'' \citep{barker:1968},
for example the kitchen while Raymond is having breakfast, the interior of
the school bus while he is travelling to school, his classroom while a
lesson is taking place, the school yard during
break.\footnote{\citet{wright:1951} record a sequence of some 1000s of
settings through which one boy progresses in a single day.}

The environmental contexts of participants in a dialogue may differ, as for
example when Mary is driving and Jack, sitting next to her, is
navigating. Here the environmental contexts of the dialogue
share in common the car interior, the road, the route ahead, and the share the same 
destination as part of their dialogue horizon. Jack's environmental context includes in addition 
the map he
is using to navigate. Mary's environmental context includes the set
of driver affordances making up the car cockpit. That dialogues of this
sort so often go wrong rests in part on the fact that there are different
ways in which space itself is demarcated in different registers
\citep{matthiessen:2014}.

Relations between environmental contexts may involve also elements of
territoriality, for example when Jack seeks to engage Mary in dialogue by
inserting himself into her personal space through displays of dominance or
enticement.  Environmental context also comprises those environments where
political or military power is projected \citep{popitz:2017}, such as the
layout of a prison in which an overseer can interact via intercom with the
prison inmates. Here the environmental context of the overseer comprehends
multiple prison security, video
surveillance and communication systems extending across the entire prison and its 
surroundings; the environmental context of the
inmate extends hardly beyond the walls of her cell.

\subsubsection{Discourse context and interpretation} \label{dc}
The dialogue \textit{is its own context} at all levels of language
production and interpretation. What this means is that, just as the
constituents of a sentence contextualise each other, so do the successive
sentences themselves. Each utterance is contextualised by its preceding utterances, and its 
potential future utterances form part of the context
horizon of each present utterance. The degree by which preceding statements
influence the interpretation of the current statement is called the
\textit{contextual weight} of these statements. In prototypical
conversations this weight decreases over time, so that the immediately
preceding utterance has the strongest weight and more remote utterances
have less as they fall away into the background. There are however cases
where interlocutors can suddenly reach back to utterances made much earlier
in the dialogue and bring them once more into the foreground. From a mathematical point of 
view, such discontinuities in the dialogue are erratic (non-Markov).

One important family of cases of this sort results from misunderstandings.
We pointed out already that acts of choosing how to respond to a
dialogue utterance are implicit. The same applies also to the
interpretations of an utterance on the part of the receiver. The latter are
observable only indirectly, for example by inference from the utterances the interpreter 
produces after a role switch between the interlocutors has occurred. This means that the 
continuous feedback which we rely on to adjust our intentions during dialogue gives us only a 
partial picture of how our interlocutor is responding to our utterances. This in turn leads to 
misunderstandings, which may remain undetected through the entire length of the dialogue. 
Where they are detected, this will often force an utterer to revise a
statement from further back in the conversation when she realises, on the
basis of how her interlocutor is now responding, that she has been
misunderstood. 

Discourse context is also present at a level above that of a single dialogue, for example when 
one dialogue is embedded inside another, or when succeeding dialogues are entangled with 
each other, as in a court case, where earlier dialogues may be inserted into the present dialogue 
context in the form of written documentation. 

\subsection{Discourse economy: implicit meaning}\label{impl}

Discourse economy occurs where the intended meaning remains partially implicit, so that the 
interpreter is required to take account of context for interpretation. Such implicit meaning is 
generated almost always unconsciously, because parties to a dialogue automatically assume 
that they share sufficient general as well as context-specific knowledge to allow each of them to
contextualise successfully the utterances of the other. Thus, they can
still effectuate an adequate interpretation, even though not everything is
said explicitly. This is of importance not least because it reflects the
way in which the structure of the dialogue is influenced by interactions
between the respective identities of its participants, above all by which intentions and 
background (linguistic and other) capabilities they 
share.\footnote{\citet[26]{verschueren:1999} gives an example of our almost universal 
reliance on dialogue economy by describing his attempt to make fully explicit the
colloquial statement: ``Go anywhere today?'' This resulted in a text of 15 lines that still does 
not achieve full.} 

The need for economy in
use of language turns on the fact that each speaker will in normal circumstances want to obtain
from her speech acts maximal effect in a limited time, and implicitness at
the right level allows her to pass over details that would otherwise
disturb the conversational flow or be boring to her interlocutor. Avoiding
explicitness can also be used as a conversational tactic, for example to
maintain politeness or mask deception.

To achieve a dialogue that is productive on both sides, the preponderance
of implicit meaning on the side of what is communicated by the utterer must still allow its 
understanding by the interpreter in a way that is close to the
utterer's intention. In his \textit{Studies in the Way of Words},
\citet{grice:1989} formulates in this connection what he calls the
``Cooperative Principle'', in which he recognises not only the need for
dialogue economy but also its two-sided nature. For cooperativeness, as
Grice understands it, incorporates both a maxim of \textit{quantity} -- be
as informative as you possibly can, and give as much information as is
needed -- and a maxim of \textit{manner} -- be as clear, as brief, and as
orderly as you can in what you say, and avoid ambiguity. These requirements
are clearly in competition with each other. If brevity is taken too far,
for example, then the interlocutors will typically later require more
explicitness in order to resolve potential misunderstandings. 

\subsubsection{Deixis}
The most important form of implicit meaning is deixis, which is the use of language elements 
whose reference is determined by some feature of the context of utterance that is in the scope 
of awareness of the dialogue partners. 
Deictic expressions -- such as ``him'',
``next week'', ``there'' -- need to be interpreted by the receiver adverting to 
features of this sort.\footnote{\citet{talmy:2018} provides a survey of such
cues as part of an account of how the utterer in a dialogue draws the
attention of the interpreter to the particular entity that she wants to
communicate about by using both speech-external and speech-internal
context. He describes the vast array of strategies humans use to bring this
about, given that the utterer cannot somehow reach into the hearer's mind
and directly place his focus of attention on that target.} Four important
forms of deixis are: person deixis, temporal deixis, spatial deixis and
discourse deixis.

\paragraph{Person deixis} \hspace{-11pt} means: references to a person, where who the
person is can be inferred only if contextual information is available
\citep{meibauer:2001, sidnell:2017}.
The utterer knows who he himself is, and in the setting of a face-to-face
communication the interpreter knows who the utterer is, and is thus able to
resolve the deictic pronouns ``I'' and ``you''.

\paragraph{Spatial deixis} \hspace{-11pt} is a phenomenon arising when reference to space
requires for disambiguation spatial features that are themselves parts of or are anchored to the 
context \citep{lyons:1977}. It can be seen at
work in the use of prepositions such as ``in'', ``out'', ``below''; also of
verbs such as ``enter'', ``go to'', ``leave''; of adverbs such as ``here'',
``there''; and of demonstrative pronouns such as ``these'' and ``those''.
For example, the utterance ``Let's go downtown'' when uttered in Berlin
needs context to be disambiguated, since ``downtown'' can mean (at least)
Berlin Zoologischer Garten and Berlin Mitte. Between 1961 and 1990 the term
``Berlin'' itself needed context for disambiguation.

\paragraph{Temporal deixis} \hspace{-11pt} is the analogous phenomenon involving reference
to time \citep{lyons:1977}. To resolve the meaning of utterances like
``Yesterday Trump met Kim'' or ``Next February I will travel to Rome''
\textit{event  time point}, \textit{time point of utterance} and
\textit{reference time scale} need to be applied in disambiguation
\citep{thomsenSmith:2018}.

The need to keep track of temporal order inside a dialogue is illustrated
by a statement such as
\begin{enumerate}
\item[(1)] After Paris we need to get to Abbeville before nightfall.
\end{enumerate}
This involves four temporal references, one (implicit) present and three
(explicit) in successive futures, as well as three spatial references:
present location at time of utterance (implicit), Paris and Abbeville
(explicit). We can use this example to illustrate how the context and
horizon of a conversation influence each other mutually. On the one hand,
if the sentence is used in a conversation between two British tourists
planning a trip from Paris to Normandy, the horizon might include potential
closing times on Somme battlefield memorial sites. If, on the other hand,
it is used in a conversation between two Oklahoma truck drivers, then the
dialogue horizon might include potential traffic holdups on Interstate 49 on the
way from Paris, Texas to Abbeville, Louisiana.

\paragraph{Discourse deixis}\label{disDeix} \hspace{-11pt} is the use of an utterance in a 
conversation to refer to this utterance itself or to previous or future parts of the
conversation \citep{levinson:1983}. Examples are: ``What you just said
contradicts your previous statements'', or ``So what does it feel like,
getting caught up in a conversation like this one?''; or again: ``This
conversation must stop immediately!'' or ``I contest the legitimacy of
these entire proceedings!'' While change in dialogue horizon normally takes
place gradually and without being noticed, the employment of discourse
deixis brings the ongoing dynamics of horizon change into the foreground.
Discourse deixis is often an element of a meta-discourse, for example when
three persons leave the room and then one of the remaining interlocutors
says: ``That was a strange conversation.''

\subsubsection{Other forms of implicit meaning}

\paragraph {Non-deictic reference} \hspace{-11pt} is a way of expressing the relation to
an entity using a fixed reference, as in proper names  or definite descriptions  
\citep{abbott:2017}. Proper names and other fixed references,
too, require background (world) knowledge to be interpreted correctly. 

\paragraph {Presupposition} \hspace{-11pt} is the usage of an implicit unit of meaning in
a way that implies that the interpreter will have to draw on contextual
knowledge to understand the intended meaning, as in the sentence ``Let us
meet the chancellor'', which carries the presupposition that the
interlocutor knows who the chancellor is. A variant type of presupposition
(as in: ``Have you stopped beating your wife?'') is sometimes used as a way
of tricking a dialogue partner in unfriendly interactions.

\paragraph{Implicature} \hspace{-11pt} occurs where there is a unit of meaning which the
speaker does not make explicit in his utterance, but which the interpreter
can deduce from this utterance. \citet{huang:2017} gives the following
example: ``The soup is warm'' implies that the soup is neither hot nor
cold. This differs from presupposition, because the implication can be
resolved without background knowledge; only minimal language competence at
the lexeme level is required.

\subsubsection{Non-interpretation: Linguistic division of labour}\label{nonInt}

Not all implicit or ambiguous lexemes or phrases have to be interpreted or disambiguated by 
every utterer or recipient of an utterance because this is not always required to realise their 
intentions. Hillary Putnam gives an important example of the interaction of utterance and 
intention in his paper on what he calls the `linguistic division of labour' 
\citep[144]{putnam:1975}. As he points out, there are many lexemes which are used by 
speakers without their full understanding. This phenomenon allows speakers and recipients to 
both tacitly use a lexeme while leaving its full understanding and definition to experts on which 
they rely, as when two politicians talk about nuclear power generation on TV. Both tacitly agree 
that they do not understand how nuclear power works, but they use the term nonetheless in 
order to sharpen their political profiles. When such tacit understanding is undermined by 
someone with genuine expertise, this leads to confusion and anger, because it adds a new, and 
undesired layer of interpretation to the dialogue in a way that disturbs their initial political 
intentions.

\subsection{Structural elements of dialogue} \label{struct}

When a human subject initiates a dialogue, she can draw, first, on multiple sets of options at 
many levels of \textit{language production}, starting with: which language to use (for
example when travelling in a foreign country); the topic to be addressed;
intonation, pitch, syntax, vocabulary, volume, as well as code and style of
language (brazen, cautious, elegant, pious, rough, wistful $\dots$); and
so on.

Second, she can draw on a wide repertoire of \textit{non-verbal
utterance accompaniments}, such as gesture, mimicry, gaze, posture. These
elements (documented in detail below) evince (or mask) underlying intentions
of the speaker, which can be argumentative, jocular, overbearing, serious,
submissive, supplicative, teasing, threatening, and so forth
(\citet{smith:2001}, section 4).

According to her intentions of the moment, the utterer can use different combinations
of the above as she adjusts to the responses of the recipient in
accordance with the physical (temporal, spatial), and social and
conversational context within which the dialogue takes place. 

The recipient of an utterance will similarly face many options on the basis of
which to attribute meaning to the utterances he hears. He can be
suspicious, trusting, fully or only partially attentive, and so on. Which
options are engaged on either side may of course change as the conversation
unfolds, either for reasons internal to the content of the conversation
itself, or because the interlocutors are influenced by external factors
such as effects of alcohol, or inclement weather, or indeed for no reason at all. Conversations 
often involve random changes of subject matter, of tone, of loudness, and so forth.

All the units that allow speakers to express their intentions are  defined as structural language 
elements \citep{verschueren:1999}. They are, from the coarse to the
fine-grained:\footnote{\citet[29f.]{ingarden:1973} identifies a similar
pattern of layers in his analysis of the ontology of the literary work of
art, and points out how each can contribute to the aesthetic quality of the
work as a whole. He emphasises that, despite the heterogeneous character of
these layers, the work nonetheless constitutes an organic unity, since the
layers are unified unproblematically by the reader in virtue of the
dimension of meaning which runs through them all. Something similar
applies in the dialogue case, though here there are two -- potentially
conflicting -- chains of meaning which unify the layers, one for each of
the two dialogue partners.}

\begin{enumerate}
\item non-verbal level: including facial expression, gestures and body language, 
\item whole language level: including language choice, language code and language style, 
\item level of single dialogue contributions: sentential and suprasentential utterance units, 
\item  level of morphemes and words,
\item  level of sound structures.
\end{enumerate}

\subsubsection{Non-verbal structural elements of dialogue} \label{nonverb}

Facial expression, glances, gestures and body language are important ways
in which uses of language are supported by non-verbal structures
\citep[100ff.]{verschueren:1999}. All of them can potentially transform
the sense of a verbal utterance, so that even a statement of condolence can
be accompanied by facial expressions that make it appear
cynical to the interpreter. In negotiations (and negotiation-based games
such as poker) body language and facial impression may be indispensable to
obtaining the desired results.
It has been shown that their effect on the interpretation of contexts,
situations and even the personality of the interlocutor is quite strong
\citep{ambady:1992}, and they form part of the deep-rooted cheater-detection mechanisms 
that have evolved in human beings to deal especially with social interactions involving 
exchange \citep{cosmides:2005}.
Communicating via gestures is also an important non-verbal component of dialogue, and is
used very often to disambiguate spatial from person deixis, including by
means of simple pointing \citep{sidnell:2017}.

\subsubsection{Language code and style} \label{code}
Code -- also called ``register'' -- is a matter of the language choices
systematically made by a social group, such as the inhabitants of an area
or the members of a social class or profession. Dialogue participants can
switch codes, for example, to convey special meaning or emphasis, or to
communicate mockery or disdain.

Style concerns the level of \textit{formality} of language use
\citep{verschueren:1999}; a speaker may switch, for example, to a more
aggressive style, in order to intimidate or punish his dialogue partner.
Both code and style are important dimensions of variance in utterance
formation and interpretation.

\subsubsection{Sentential and suprasentential utterances} 
A sentential utterance expresses a relatively closed unit of meaning
encompassing the basic functions of reference and predication. Subkinds
are: statement, question, command, request and exclamation
\citep{duerscheid:2012}. Statements are characterised by features such as
reference (subject in noun phrase) and predication (verb phrase).
Typically, they are expressed as complete sentences, but ellipses are also
used, as in ``Guilty, your honour''. Such expressions are also a form of  
dialogue economy.

A suprasentential utterance is a sequence of sentential utterances which the utterer uses to 
optimise the fulfilment of her intentions by conveying her meaning in corresponding detail. 
The way this is done, too, depends on context.

\paragraph{Incompleteness and ellipses}\label{ellipse}
Sentential and suprasentential utterances are often incomplete or
elliptical. This may result from interruption or from the inability of the
speaker to finish his thought. But often, such utterances can be completed
by elements of the situation and are not pragmatically incomplete
\citep{mulligan:1997}. In such cases humans can interpret even incomplete
utterances in a sense that is close to the meaning intended by the utterer.

\paragraph{Force and modality}
Force describes utterance styles characteristic of assertion, command,
request, question, and so forth. In addition, there are varying degrees of
force, so that, depending on the emotional involvement and inclinations of
the speaker, a request to obtain something might be phrased either as a
question or as an imperative.

With \citet{frege:1879} and \citet{searle:1978} one might take the view
that an expressed proposition can be evaluated independently of the force
with which it is communicated. \citet{hanks:2007}, however, gives strong
evidence to the effect that propositional content and force interact. Thus, while logicians and 
computer scientists
have sometimes held that the linguistic subdiscipline of semantics can hold
itself separate from concerns with matters of pragmatics, a view of
this sort cannot be maintained even for the language used in silent
monologue \citep{clark:1996}. Such a view will certainly be inadequate when
it comes to that sort of language whose mastery is needed to conduct a convincing 
conversation.

The philosopher's understanding of force is closely related to the
linguistic notion of \textit{modality}, which describes aspects of attitude
-- of how the utterer relates to his utterance, signalling properties such
as: degree of certainty, optionality, urgency, hesitancy, vagueness, possibility,
necessity, and so forth \citep{verschueren:1999}.

But modality as understood by linguists comprehends also other aspects of
the utterer's attitude, for example that he is joking, lying, flattering,
ordering, arguing, interrogating, pleading. The verbal expression of
modality is often combined with non-verbal language-supporting elements
(see \ref{nonverb}), for example when the utterer is holding a gun to the
head of the interpreter, or is kneeling before the interpreter in the
middle of the street while holding a ring in his hand.

\paragraph{Lying and deception} 

Lying and deception are frequent phenomena in language use; \citet{nietzsche:1873} even sees 
them as an essential part of language usage. Their source is the desire to achieve one's goals 
and intentions without the knowledge of the interlocutor or against her will. Lying changes the 
entire meaning of a dialogue both for the deceiver and, in the case that she becomes aware of 
the deception, for the deceived person. Sometimes the deception may be made explicit by the 
deceived person (``You must be lying to me because at that time you could not have been at 
home!''), but otherwise it remains implicit because the deceiver will have no motive to reveal it. 
It is generally therefore not possible to model lying and deception using what is observable in a 
dialogue. 

\subsubsection{Lexemes} 
Lexemes are the carriers of the minimal units of linguistic meaning -- for
example \textsc{run} or \textsc{hat}. The building blocks of sentences are
lexemes in their inflected forms, which are called wordforms -- for example
\textit{runs}, \textit{ran}, \textit{running} or \textit{hats},
\textit{hat's}, \textit{behatted}. For any given language there is a
relatively small set of lexemes that has to cover a very wide range of
possible topics. This is because it is not possible to have an exact word
for each and every aspect of reality if the size of the lexicon is to be
kept small enough that it can be managed by a single human being. Lexemes
are therefore prototypes \citep{rosch:1975}. They obtain part of their
meaning from the context created by the other lexemes they are used with in
a sentence, as well as by all the other contextual dimensions identified above.
For example, the lexeme \textsc{freedom} has a very different meaning in (2) and (3):

\begin{enumerate}
\item[(2)] Do not clutter my desk with stuff, I need freedom to move.
\item[(3)] We want freedom of speech!
\end{enumerate}

Depending on intention and context, lexemes at varying levels of abstractness and generality
may be chosen in the course of a single dialogue. In everyday
usage it is the mid-level that dominates. For instance, when talking about
pets, the participants in a dialogue will typically use mid-level terms
such as ``dog'' or ``cat'' rather than the low-level ``dachshund'' or the
high-level ``animal''. Something similar holds when we describe an ailment
(where we refer in a dialogue to a fracture of the \textit{foot}, rather
than of the \textit{fifth metatarsal bone}). When we introduce ourselves in
a dialogue, we (prototypically) talk about our place of origin by referring
to city or region rather than to neighbourhood or street. Utterers,
normally unconsciously, select the level of abstractness or generality that is
salient to the dialogue context (compare \ref{dc}).

\subsubsection{Sound structure} \label{soundS}
In its sound structure, speech is built out of elementary phonetic segments
(vowels and consonants), which are combined into composite sounds beginning
with syllables and words and proceeding to entire sentential and
supersentential utterances. We can compare the former to single notes in
music, and the latter to melodic structures formed by notes in sequential
combination.\footnote{The elementary phonetic segments of vocal utterances
have features comparable to the pitch, overtone composition, and amplitude
of single notes in music.} Each entire utterance is made with a specific
\textit{prosody}, by which is meant that aspect of speech sound that
inheres in composite sound units. Among the various dimensions of prosody,
intonation and pace are the most important.

Variations in \textit{intonation} -- for example suddenly switching to a
high-pitched voice -- are used to express emotions or attitudes of the
speaker, or to distinguish sentential units of different modalities (for
example statements from questions), or for purposes of emphasizing or
highlighting certain aspects of the dialogue, or to regulate the
conversational flow.

Sounds and the meanings they convey are tightly associated with our physical experience, as
is evident from visceral reactions such as nausea or sexual excitement that certain speech 
sound types can evoke.\footnote{Note that the pitch variation in intonation
is different from \textit{tone}, another type of pitch modulation, that is
used to distinguish  grammatical or lexical meaning. In Mandarin, for
example, lexemes are differentiated via differences in what is called
syllable pitch.} Another aspect of sound structure that can influence
interpretation is \textit{voice quality}, such as the use of a soft or hard
voice, or the use of mere vocal cues such as throat-clearing, grunts,
sniffs, unintelligibly muttering under one's breath.\footnote{We focus here
on sound structure as it appears in the flow of a spoken dialogue. But
sound structure can play a role, too, in written dialogue, for example when
our minds associate the words in an email message with a certain
intonation. This is an example of the subtlety and massive complexity of
language interpretation as it occurs in the dynamic flow of inner mental
experience.}

\textit{Pace} comprises rhythm, speed -- for example speeding up or pausing,
hesitating in mid-sentence -- all of which can be selected, consciously or
unconsciously, to shape the ways an utterance or sequence of utterances is
interpreted. Pausing can also be used as a device to signal to one's
interlocutor that a conversation is reaching its end. Different types and
layers of sound can be used together, for example when a dialogue partner responds to an 
utterance with a slow hand clap, or when Romeo serenades his sweetheart with musical
accompaniment.

\subsection{Dialogue dynamics} \label{dyn} 

The production of meaning in the course of a dialogue is a highly dynamic
process, which may unfold on all of the levels distinguished above: the
intentions of the interlocutors, the dialogue horizon generated by the
interaction of relevant dialogue contexts (see \ref{context}), deixis and
other forms of implicit meaning, and all the dialogue's structural
elements. As we have seen above, the ways dialogue participants interpret each other's 
utterances depends on their past experiences, and may have a strong emotional component  
\citep{drace:2013}. In their utterance production, each can draw on a huge variety of
interacting combinations of the structural elements. Moreover, while this is happening, the 
dialogue horizon itself is evolving: some things and processes move into the field of what is 
relevant to the dialogue, others fall away. Where the dialogue itself becomes its own context, 
this leads to a refocussing and potentially to a reinterpretation of all earlier contexts, which 
then influences how subsequent (unconscious
and conscious) choices will be made in utterance formation and interpretation as the 
conversation proceeds. For example, an interlocutor might say: "The facts that you bring up 
now contradict the conclusions you drew just half an hour ago." 

\subsubsection{Dialogue flow interruptions}

While in a prototypical dialogue the interlocutors take turns, clean and regular turn taking is 
rather exception than rule, because most dialogues contain interruptions, and sometimes, as in 
the conversational style favoured among Parisian intellectuals, consists entirely of 
interruptions. Conversational turn-taking is displayed in its ideal form in the strings of 
characters printed by a teleprinter on a moving paper tape, where only one
person can have control over the input mechanism at any one time. This
ideal form is illustrated also by a published interview after an editor has
worked to create a polished textual flow.

In actually occurring spoken
dialogues, however, there are frequent deviations from this
ideal. The utterer may pause or hesitate or stutter, create false starts,
make mistakes, interrupt herself or try to add retrospective corrections to
what she has said earlier, suddenly change the subject of the dialogue
entirely. The interpreter may seize the speaker role by forcing a role
switch before the utterer has finished her statement. If the utterer does
not yield to the interruption, this leads to utterances occurring
simultaneously, so that the flow of meaning transmission breaks. Sometimes,
the interpreter anticipates the next statement of the utterer and takes a
turn before the latter has finished. All these deviations increase the
complexity of the role context and add to the pressures on the dialogue
participants both in forming and in interpreting dialogue utterances. They
often go hand in hand with emotional layers to the dialogue flow, which
support specific sorts of interpretation of dialogue utterances, for
example where one dialogue partner seeks to influence the other by (as we
say) playing on his emotions.

\subsection{Summary remarks on dialogue variance}

We invite the reader to note not merely the many \textit{levels of dialogue variance} 
distinguished in the above but also the
degree to which these variations depend on multiple factors (indeed
multiple levels of multiple factors), both inside and outside the dialogue
itself, factors which can extend to include almost any matter within the
biographies and within the scope of the knowledge and interests of the
dialogue partners.

We note further the degree to which many of these factors are a matter of
continuous variation in the sense that the range of options forms a
continuum, as for example between speaking with a soft and a loud voice, or
with a calm and an angry voice. 

Movements along multiple such continua may
take place within a single dialogue, and when such movements are effected
by one dialogue partner they will typically call forth some concordant
movement on the side of her interlocutor.

In all respects, indeed, preserving the flow of a dialogue rests on the
capacity of humans to adjust their contributions to fit those of the
dialogue partner, for example to adjust their respective intentions. 

This capacity is applied even in the most heated of disputes between friends or
lovers, where even the most acrimonious of dialogue partners are able to
maintain a conversation flow for considerable periods of time. This is
achieved through a type of homeostatic process, whereby, when the
conversation seems to be going completely off the rails, one or other
partner succeeds in pulling it back from the brink and initiating another
phase of what is once more recognizable as coherent verbal exchange.

\subsection*{Acknowledgements}
For comments on an earlier version of this manuscript we would like to thank Larry
Hunter, Prodromos Kolyvakis, Niels Linnemann, Emanuele Martinelli, Robert Michels, Alan Ruttenberg, and Thomas Weidhaas.
 \printbibliography
\end{document}